\documentclass[conference]{IEEEtran}
\usepackage{cite}
\ifCLASSINFOpdf
   \usepackage[pdftex]{graphicx}
\else
\fi

\usepackage{amsmath}
\usepackage{mathtools}

\usepackage{stfloats}
	
\usepackage{xcolor}
\usepackage{booktabs}
\usepackage{url}
 
\usepackage{url}

\usepackage[colorlinks,urlcolor=blue]{hyperref}

\DeclareMathOperator*{\argmin}{arg\,min}
\begin{document}
%
\title{Forecasting Player Behavioral Data \\ and Simulating in-Game Events}
%
 \author{\IEEEauthorblockN{Anna Guitart, Pei Pei Chen, Paul Bertens and \'{A}frica Peri\'a\~{n}ez}
 \IEEEauthorblockA{
 Game Data Science Department\\
 Silicon Studio\\
 1-21-3 Ebisu Shibuya-ku, Tokyo, Japan\\
 \{anna.guitart, peipei.chen, paul.bertens, africa.perianez\}@siliconstudio.co.jp}
 }

\maketitle

\begin{abstract}
Understanding player behavior is fundamental in game data science. Video games evolve as players interact with the game, so being able to foresee player experience would help to ensure a successful game development. In particular, game developers need to evaluate beforehand the impact of in-game events. Simulation optimization of these events is crucial to increase player engagement and maximize monetization.
We present an experimental analysis of several methods to forecast game-related variables, with two main aims: to obtain accurate predictions of in-app purchases and playtime in an operational production environment, and to perform simulations of in-game events in order to maximize sales and playtime. Our ultimate purpose is to take a step towards the data-driven development of games. 
The results suggest that, even though the performance of traditional approaches such as ARIMA is still better, the outcomes of state-of-the-art techniques like deep learning are promising. Deep learning comes up as a well-suited general model that could be used to forecast a variety of time series with different dynamic behaviors. 
\end{abstract}


%
\IEEEpeerreviewmaketitle

\section{Introduction}
In the last few years, we have witnessed a genuine paradigm change in the development of video games \cite{el2013game,yannakakis2017ai}. Nowadays, petabytes of player data are available, as every action, in-app purchase, guild conversation or in-game social interaction performed by the players is recorded. This provides data scientists and researchers with plenty of possibilities to construct sophisticated and reliable models to understand and predict player behavior and game dynamics.
Game data are time-dependent observations, as players are constantly interacting with the game. Therefore, it is paramount to understand and model player actions taking into account this temporal dimension.

In-game events are drivers of player engagement. They influence player behavior due to their limited duration, strongly contributing to the in-game monetization (for instance, an event that offers a unique reward could serve to trigger in-app purchases). Those events break the monotony of the game, and thus it is essential to have a variety of types, such as battles, rewards, polls, etc. Anticipating the most adequate sequence of events and the right time to present them is a determinant factor to improve monetization and player engagement.

\begin{figure}[hb!]
  \centering
  \includegraphics[width=0.49\textwidth]{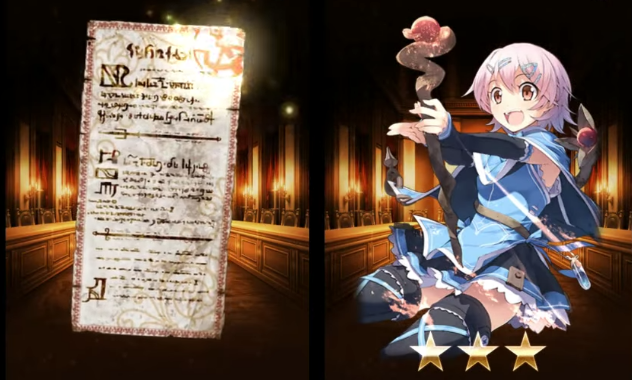}  
\caption{Screenshot of an in-game gacha event in \emph{Grand Sphere} developed by \emph{Silicon Studio}. The left panel shows an item that players can purchase, which on opening reveals an in-game card (shown in the right panel), in this case having 3 stars. The card obtained is random, and cards with more stars are more valuable and also rarer. Different in-game events can modify the probability of getting cards with more stars or the types of cards that can be obtained. }
\label{gacha}
\end{figure}


Forecasting time series data is a challenge common to multiple domains, e.g. weather or stock market prediction. Time series research has received significant contributions to improve the accuracy and time horizon of the forecasts, and an assortment of statistical learning and machine learning techniques have been developed or adapted to perform robust time series forecasting \cite{de200625,brockwell2016introduction}.

Time series analysis focuses on studying the structure of the relationships between time-dependent features, aiming to find mathematical expressions to represent these relationships. Based on them, several outcomes can be forecast \cite{adhikari2013introductory}. 

Stochastic simulation \cite{asmussen2007stochastic} optimization consists on finding a local optimum for a response function whose values are not known analytically but can be inferred.
On the other hand, the analysis of \emph{what-if} scenarios (or just simulation optimization) is the process of finding the best inputs among all possibilities that maximize an outcome \cite{carson1997simulation}.

The aim of this work  is twofold: on the one hand, to accurately forecast time series of in-game sales and playtime; on the other, to simulate events in order to find the best combination of in-game events and the optimal time to publish them. To achieve these goals, we performed an experimental analysis utilizing several techniques such as ARIMA (dynamic regression), gradient boosting, generalized additive models and deep neural networks \cite{box1976time,friedman2001greedy,hastie1990generalized,busseti2012deep}.  

Pioneering studies on game data science in the field of video games, such as \cite{bauckhage2012players,hadiji2014predicting,perianez2016churn,GameBigData}, concentrated in churn prediction. 
Other related articles that analyze temporal data in the game domain   \cite{bauckhage2015clustering,drachen2012guns,drachen2014comparison,sifa2014playtime,saas2016discovering} focused on unsupervised clustering, not in supervised time series forecast. To the best of our knowledge this is the first work in which forecasts of time series of game data and simulations of in-game events are performed.

%

%
  \section{Model Description}
  \subsection{Autoregressive Integrated Moving Average (ARIMA)}

ARIMA was firstly introduced by Box and Jenkins \cite{boxJenkins} in 1976, in a book that had a tremendous impact on the forecasting community. Since then, this method has been applied in a large variety of fields, and it remains a robust model, used in multiple operating forecast systems \cite{lawrence2006advances}. ARIMA characterizes time series focusing on three aspects: 1) The autoregressive (AR) terms model the past information of the time series, 2) The integrated (I) terms model the differencing needed for the time series to become stationary, e.g. the trend of the time series, 3) The moving average (MA) terms control the past noise of the original time series. 

Specifically, the AR part represents the time series as a function of $p$ past observations,
\begin{equation}
y_{t} = \varphi_1 y_{t-1} + \cdots + \varphi_p y_{t-p},
\end{equation}
with $\varphi_1,\ldots,\varphi_p$ the AR coefficients and $p$ the number of past observations needed to perform a forecast at the current time.
The MA component, rather than focusing on past observations, uses a moving average of past error values in a regression model, 
\begin{equation}
y_{t} = \varepsilon_{t} + \theta_1 \varepsilon_{t-1} + \cdots + \theta_q \varepsilon_{t-q},
\end{equation}
where $q$ is the number of moving average lags, $\theta_1,\ldots,\theta_q$ are the MA coefficients and $\varepsilon_t,\ldots,\varepsilon_{t-q}$ the errors. 
Finally,  a $d$ parameter is used to model the order of differencing, i.e. the I term of ARIMA:
\begin{equation}
y^*_{t} = y_t-y_{t-1} - \cdots -  y_{t-d}.
\end{equation}
Taking $d=1$ is normally enough in most cases \cite{adhikari2013introductory}.
If $d=0$, the model is reduced to an ARMA$(p,q)$ model.

The ARIMA model can only analyze stationary time series; however most of the stochastic processes exhibit non-stationary behavior. Only through the differencing operation, or by applying a previous additional transformation to the time series, e.g. a log or Box--Cox \cite{boxCox} transformation, we can convert the time series into stationary objects.   

We can write the ARIMA model as
\begin{equation}
\varphi(L)(1-L)^d y_t = \theta(L) \varepsilon_t,
\end{equation}
where $L$ is the lag operator, i.e. $L\varphi_t=\varphi_{t-1}$. Therefore, we can express the ARIMA model in terms of the $p$, $d$ and $q$ parameters as
\begin{equation}
\bigg(1 - \sum^{p}_{i=1} \varphi_i L^i \bigg)  \big(1-L\big)^d y_t = \bigg(1 +  \sum^{q}_{j=1} \theta_j L^j \bigg) \varepsilon_t.
\label{ARIMAformula}
\end{equation}
Box and Jenkins also generalized the model to recognize the seasonality pattern of the time series by using $y_t = {\rm ARIMA}(p,d,q)(P,D,Q)_m$ \cite{boxJenkins}. The parameters $P$, $D$, $Q$ represent the number of seasonal autoregressive, differencing and moving average terms. The order of the seasonality (e.g. weekly, monthly, etc.) is set by $m$. 
In order to determine the ARIMA parameters, the autocorrelation function (ACF) and partial autocorrelation function (PACF) are analyzed \cite{boxJenkins}. Once the parameters are fixed, the model is fitted by maximum likelihood and selected (among the different estimated models) based on the Akaike (AIC) \cite{akaikeAIC} and Schwarz Bayesian (BIC) \cite{schwarzBIC} information criteria. 

\subsubsection*{Dynamic Regression}
\label{dynamic}
To include the serial dependency of external covariates, a multivariate extension of Equation \ref{ARIMAformula} is proposed, the so-called \emph{dynamic regression} (DR). The relationship between the output, its lags and the variables is linear.  
DR takes the form \cite{cragg1982estimation}

\begin{multline}
\bigg(1 - \sum^{p}_{i=1} \varphi_i L^i \bigg) \big(1-L\big)^d y_t \\
=\sum^{k}_{r=1}\gamma_{r} \bigg(1 - \sum^{p}_{l=1}  \varphi_l L^l \bigg)\big(1-L\big)^d x_{rt}\\  +  \bigg(1 + \sum^{q}_{j=1} \theta_j L^j \bigg) \varepsilon_t.
\label{ARIMAregression}
\end{multline}

The first term represents the AR and I parts, the last corresponds to the MA component and the middle term is where the $k$ external variables $x_t$ are included. The $\gamma_{r}$ are the corresponding coefficients of the covariates.

\begin{figure*}[ht!]
  \centering
  \includegraphics[width=\textwidth]{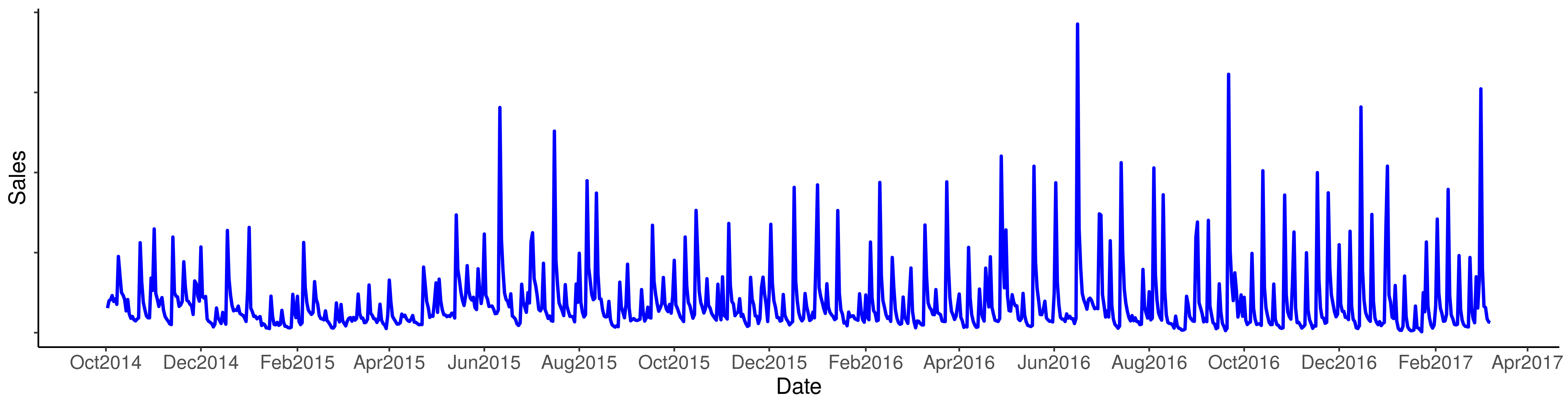}
  \includegraphics[width=\textwidth]{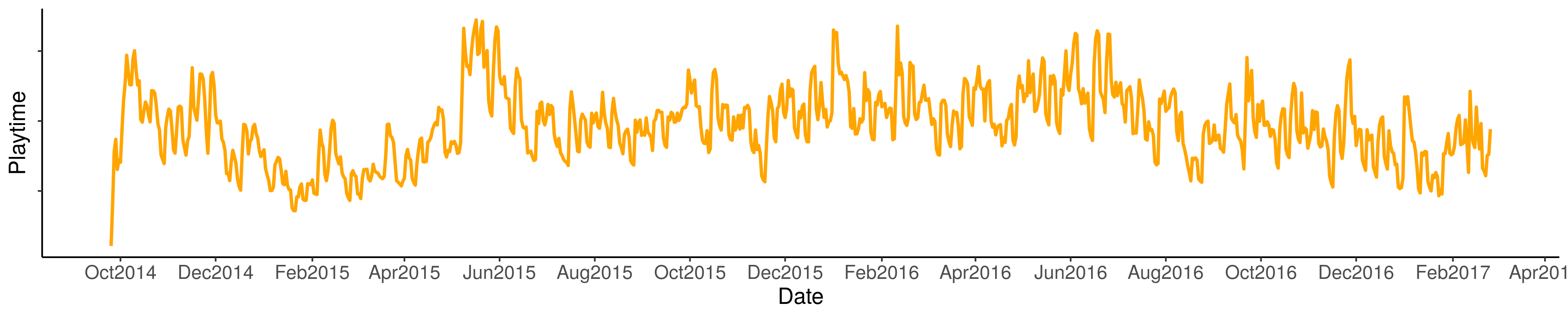}
\caption{Time series of the total daily sales (above) and total daily playtime (below) from \emph{Age of Ishtaria} over a period of ${\sim}2.5$ years. In this and other figures below, the quantitative sales and playtime values are not shown in the vertical axes for privacy reasons.}
\label{data_Ishtaria}
\end{figure*}

\subsection{Gradient Boosting Models}
Gradient boosting machines (GBMs) \cite{dietterich2000ensemble} are ensemble-based machine learning methods capable of solving classification and regression problems\cite{friedman2001greedy}. A GBM consists of an ensemble of weak learners (i.e. learners that perform only slightly better than random classifiers), commonly decision trees. Each weak learner is sequentially added to the ensemble, thus continuously improving its accuracy \cite{mason1999boosting}. The approach taken by GBMs  stems from the idea that boosting (using multiple weak models to create a strong model) can be seen as an optimization algorithm on some suitable loss function\cite{breiman1997arcing}. If this loss function is differentiable, the gradient can be calculated and gradient descent can be used for optimization \cite{friedman2001greedy}. This makes it possible to recursively add weak learners that follow the gradient, minimizing the loss function and reducing the error at each step. The construction of an ensemble to fit a desired function $f(x)$ is illustrated below (a more in-depth analysis can be found in \cite{ridgeway2007generalized}). For each weak learner in the ensemble, we sequentially do the following: First, we compute the negative gradient


\begin{equation}
z_{i}=-\frac{\delta}{\delta f(x_i)}\mathcal{L} (y_i,f(x)) |_{f(x_i)=\hat{f}(x_i)},
\end{equation}
where $\mathcal{L}$ is the loss function and $i$ the current weak model.

Then, we fit a regression model $g(x)$ predicting $z_i$ from $x_i$ and apply gradient descent, with step size given by
\begin{equation}
\rho (x) = \argmin_\rho \sum^N_{i=1} \mathcal{L} (y_i,f(x_i)) + \rho g(x_i).
\end{equation}
Finally, we update the estimate of $f(x)$ through 
\begin{equation}
\hat{f}(x)\leftarrow\hat{f}(x)+\rho g(x).
\end{equation}


GBMs can effectively capture complex non-linear dependencies \cite{natekin2013gradient} and several strategies to avoid overfitting can be applied, e.g. shrinkage \cite{friedman2001greedy} or early stopping \cite{zhang2005boosting}.


\subsection{Generalized Additive and Generalized Additive Mixed Models}
The generalized additive models (GAMs) derived by \cite{hastie1987generalized} are a combination of generalized linear (GLMs) and additive (AMs) models. In this way, GAMs exhibit the properties of both, namely the flexibility to adapt to any distribution of GLMs and  the non-parametric nature of AMs.

The structure of a GAM is
\begin{equation}
	g(E(y)) = \beta_0 + s_1(x_1) + \cdots + s_n(x_n),
\end{equation}
where $n$ is the number of predictors, $E(y)$ is the expected value, $g(\cdot)$ is a link function, and $s_i(\cdot)$ are smooth functions.

Even though the distribution of the response variable is the same as for GLMs, there is a generalization that allows GAMs to accommodate different kinds of responses, for example binary or continuous. GAMs assume that the means of the predictors $x_i$ are related to an additive response $y$ through a nonlinear link function $g$ (such as the identity or logarithm function). The model distribution can be selected from e.g. a Gaussian or Poisson distribution \cite{hastie1990generalized}.

As additive models, in contrast to parametric regression analysis (which assumes a linear relation between responses and predictors), GAMs serve to explore non-parametric relationships, as they make no assumptions about those relations. Instead of using the sum of the individual effects of the predictors as observations, GAMs employ the sum of their smooth functions, which are called \emph{splines} \cite{maindonald2010smoothing} and include a parameter that controls the smoothness of the curve to prevent overfitting. With the link and smooth functions mentioned above, the GAM approach has the flexibility to interpret and regularize both linear and nonlinear effects \cite{larsengam}. 

However, GAMs do not assume correlations between observations (such as the time series temporal correlation in this study). As a consequence, when these are present, another approach would be better suited to perform the forecast. Generalized additive mixed models (GAMMs) \cite{chen2000generalized}, an additive extension of generalized linear mixed models (GLMMs) \cite{breslow1993approximate}, constitute such an approach. GAMMs incorporate random effects $u$ and covariate effects $\gamma_j$ (which model the correlations between observations by taking the order of observations into account) into the previous GAM formulation. When estimating the $j$th observation, the structure of a GAMM is 

\begin{equation}
	g(E(y_j)) = \beta_0 + s_i(x_{ji}) + \gamma^\top_j u.
\end{equation}

Smoothness selection can be automatized \cite{wood2011fast} and the estimation of the GAMM is conducted via a penalized likelihood approach.

\subsection{Deep Belief Networks}

Deep neural networks (DNNs)\cite{bengio2009learning} have been used with great success in recent years, achieving cutting-edge results in a wide range of fields \cite{deng2014deep}.  This method has outperformed alternative models on almost all tasks from image classification\cite{krizhevsky2012imagenet} to speech recognition\cite{graves2013speech} and natural language modeling\cite{deng2014deep}.

Deep belief networks (DBNs)\cite{hinton2006fast} are an extension of DNNs where the units in the hidden layers are binary and stochastic. They are capable of learning a joint probability distribution between the input and the hidden layers and can be used as a generative model to draw samples from the learned distribution. A DBN uses a stack of restricted Boltzman machines (RBMs)\cite{ackley1985learning}, where each RBM is trained on the result of the previous layer as follows:
\begin{gather}
   P(h_i^{(k)}=1|h^{(k+1)})=\sigma\bigg(b_i^{(k)}+\sum_jW_{ij}^{(k)}h_j^{(k+1)}\bigg),
\end{gather}
with $W^{(k)}$ being the weight matrix between the layers $k$ and $k+1$ and $\sigma$ being the logistic sigmoid.  

For supervised learning, a DBN can be used as a way to pre-train a DNN \cite{larochelle2008classification}. Each layer of the RBM is first trained separately, and then the weights in all layers are fine-tuned through standard back-propagation. 
A more comprehensive overview of DBNs and DNNs can be found in \cite{hinton2006fast} and \cite{bengio2009learning}. 

There have been previous works on applying DNNs to time series forecasting problems (see e.g. \cite{langkvist2014review} and \cite{busseti2012deep}) and comparisons between ARIMA and artificial neural networks (ANNs) have also been performed \cite{zhang1998forecasting}. However, when dealing with small datasets, standard DNNs can quickly overfit due to having too many parameters. Careful regularization is required to ensure that the model can be generalized to unseen data. Recent techniques like drop-out (which randomly masks some of the hidden units) \cite{srivastava2014dropout} or stochastic DBNs can significantly help with this problem, as they constrain the number of parameters that can be modified. Additionally, using $L_2$ regularization on the weights\cite{ng2004feature} and ensuring proper selection of the number of hidden units can make DBNs an effective model even when faced with smaller datasets.   

\begin{figure}[h!]
  \centering
  \includegraphics[width=0.49\textwidth]{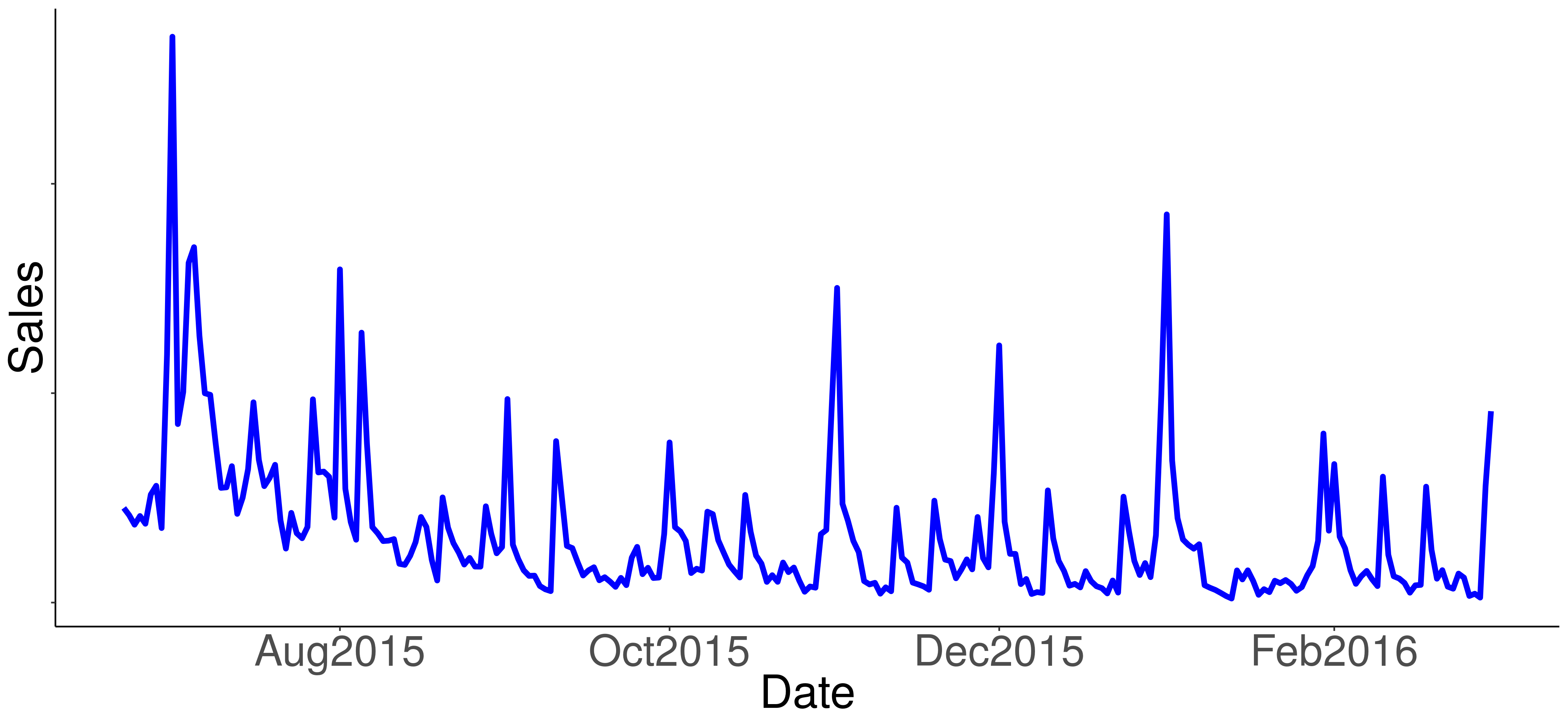}
  \includegraphics[width=0.49\textwidth]{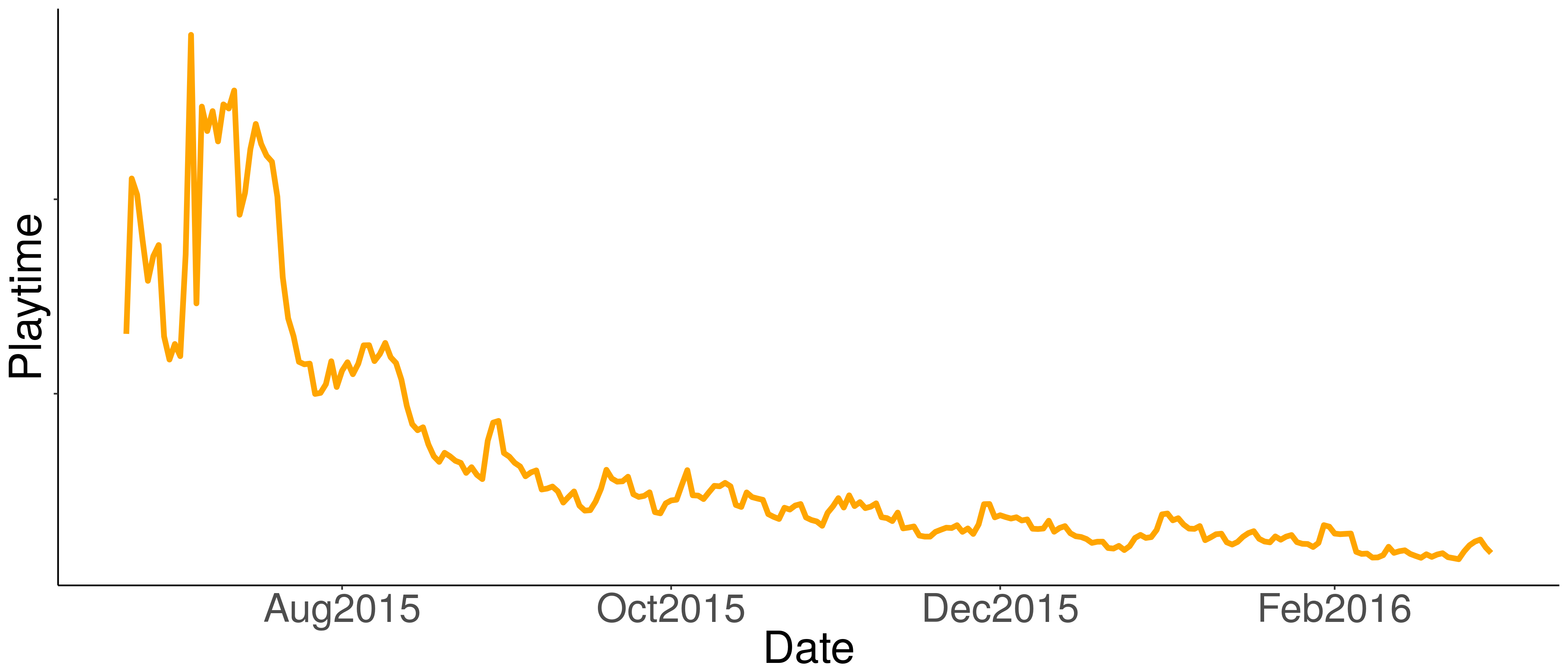}
\caption{\emph{Grand Sphere} time series of total daily sales (above) and total daily playtime (below) over a period of ${\sim}9$ months. Quantitative values are not shown in the vertical axes for privacy reasons.}
\label{data_Grand Sphere}
\end{figure}

\section{Forecast Performance Metrics}
\label{errorMetric}
In order to validate the accuracy of forecasting models, several performance measures have been proposed in the literature \cite{MASE}. 
The use of various metrics to analyze time series is a common practice. The ones selected in this study possess different properties, which is important to correctly assess the forecasting capabilities of the models from different perspectives (e.g. in terms of the magnitude or direction of the error).
Each of the metrics summarized here is a function of the actual time series and forecast results. In all the equations below, $n$ refers to the number of observations, $f_i$ denotes the forecast value and $a_i$ represents the actual value. 

\subsection{Root Mean Squared Logarithmic Error (RMSLE)}
The RMSLE can be defined as
\begin{equation}
	\text{RMSLE} = \sqrt{\frac{1}{n}\sum_{i=1}^{n}\Big(\log(f_i + 1) - \log(a_i + 1)\Big)^2}.
\end{equation}
Because the RMSLE uses a logarithmic scale, it is less sensitive to outliers than the standard root mean square error (RMSE). 
Additionally, the RMSE has the same tendency to underestimate and overestimate values, whereas the RMSLE penalizes more the underestimated predictions.

\subsection{Mean Absolute Scaled Error (MASE)}
The MASE is a scale independent metric, i.e. it can be used to compare the relative accuracy of several forecasting models applied to different datasets. To calculate this metric, errors are scaled with the in-sample mean absolute error (MAE). The exporession for the MASE is
\begin{equation}
	\text{MASE} = \frac{1}{n}\sum_{i=1}^{n}\left(\frac{\vert f_i - a_i \vert}{\frac{1}{n-1}\sum_{i=2}^{n}\vert a_i - a_{i-1} \vert}\right).
\end{equation}

\subsection{Mean Absolute Percentage Error (MAPE)}
The MAPE is estimated by
\begin{equation}
	\text{MAPE} = \frac{100}{n}\sum_{i=1}^{n}\Big\vert\frac{f_i - a_i}{a_i}\Big\vert .
\end{equation}
Since it is a percentage error measure of the average absolute error, the MAPE is also scale-independent. However, if the time series have zero values, the MAPE yields undefined results because of a division by zero \cite{MASE}. The MAPE is also biased towards underestimated values and does not penalize large errors. 
  
\begin{table*}[t!]
\centering
\caption{Tuned parameters for each model}
\begin{tabular}{clcccccccccccccccc} \toprule
&&\multicolumn{2}{c}{} & \multicolumn{7}{c}{Models}\\ \cmidrule(l){3-17}  
&& \multicolumn{2}{c}{GBM}  &  \multicolumn{7}{c}{ARIMA} &  \multicolumn{6}{c}{DBN} \\ \cmidrule(lr){3-4}\cmidrule(lr){5-11}\cmidrule(l){12-17}
\multicolumn{2}{c}{Dataset}     & max\_depth & eta  &  p & d & q & P & D & Q & m & h & n & plr & tlr & k & b \\ \midrule 
Sales & Age of Ishtaria    &   100  & 0.20 & 2 & 1 & 1 & 1 & 1 & 1 & 7 & 2 & 50 & 0.0001 & 0.01 & 5 & 50 \\ 
Sales & Grand Sphere       &   1    & 0.76 & 2 & 1 & 1 & 1 & 0 & 1 & 7 & 2 & 300& 0.001  & 0.1  & 2 & 10 \\
Playtime & Age of Ishtaria &   1    & 0.66 & 2 & 1 & 2 & 1 & 1 & 1 & 7 & 2 & 50 & 0.0001 & 0.01 & 2 & 50\\
Playtime & Grand Sphere    &   1000 & 0.23 & 1 & 1 & 1 & 1 & 1 & 1 & 7 & 2 & 300& 0.0001 & 0.1  & 2 & 50 \\ \bottomrule
\end{tabular}
\label{modelParameters}
\end{table*}

\section{Dataset}
\subsection{Data Source}
 The study presented in this article focuses on the analysis of two different daily time series, those of \emph{playtime} and \emph{sales}. The data was collected from two Japanese game titles developed by Silicon Studio: \emph{Age of Ishtaria} (hereafter, \emph{AoI}) and \emph{Grand Sphere} (hereafter, \emph{GS}). Playtime corresponds to the total amount of time spent in the game by all users, while sales represents the total amount of in-app purchases. For \emph{AoI}, daily information was extracted from October 2014 until February 2017 (${\sim}2.5$ years) and for \emph{GS}, from July 2015 until March 2016 (${\sim}9$ months). Additionally, data from all the game events, marketing and promotion campaigns within the collection period were also gathered to be used as external variables for the model. Figures \ref{data_Ishtaria} and \ref{data_Grand Sphere} show the two daily time series for \emph{AoI} and \emph{GS}, respectively, which present clear differences concerning trends and seasonal behavior.

\subsection{External features}
\label{external}

Proper identification and subsequent removal of outliers caused by unexplained phenomena can significantly improve the modeling of the time series \cite{outliers1972}. To that end, anomaly detection using a deep autoencoder \cite{sakurada2014anomaly} was performed. This technique is capable of finding subtler anomalous behaviors than traditional methods, and shows that the outliers coincide with the external events of the game that take place on the same particular day. These events are derived from in-game information and included in the model as external variables. They can be:
\begin{itemize}
	\item \emph{Game Events}: Events such as raid battles, boss battles, etc. Each type of event is input separately.
	\item \emph{Gacha}: A monetization technique used in many successful Japanese free-to-play games. It describes an event where players can spend money to randomly pull an item from a large pool (inspired by capsule toy vending machines).
	\item \emph{Promotions}: Release of new cards and discounts to engage current users.
	\item \emph{Marketing Campaigns}: Acquisition campaigns launched to obtain new users.
\end{itemize}
In the case of \emph{gacha} and \emph{promotions}, the corresponding \emph{event scale} was also taken into account. This scale is used to quantify the outlier effect. It represents the influence or importance of the event in question and is related to the amount of money invested in it. Specifically, the event is assigned a value 1, 2, 3 or 4 to denote low, medium, high or super-high influence. 


Other non-game-related features included in the model are:
\begin{itemize}
    \item \emph{National Holidays}: National Holidays in Japan. \cite{holidaysTokyo} 
    \item \emph{Temperature}: Average daily temperature in Tokyo \cite{tempTokyo}. 
\end{itemize}


Moreover, the day of the week and month of the year were added as extra input for the GAMM and GBM, to make it possible for the model to learn the seasonality effects. For the ARIMA model these data do not need to be included as they are already inherently considered by the seasonal parameters.

\section{Method}

\subsection{Data preparation}

To make the original time series stationary, a \emph{Box--Cox} transformation \cite{box1964analysis} is applied to the sales data, and a \emph{logarithm} transformation to the playtime data, for the DR, GBM and DBN models. The GAMM is the only technique that does not require a prior transformation of the data.

The categorical values for the external features (game events, \emph{gacha}, promotions, marketing campaigns, national holidays, etc.) are included in the model as step functions (dummy variables). For each day in the training and forecasting data, the covariates are encoded with a vector that is either 0 or 1 depending on the absence or presence of the event on that date. However, for events with an \emph{event scale}, the vector value matches the corresponding scale value instead.

Since promotions and marketing campaigns can have some delayed effects on the series, input is added to the days after the campaign release by means of a decay function. For marketing campaigns, one-week effects are considered, and their values are assumed to decrease linearly with time; on the other hand, for promotions, the decrease is dependent on the scale of the campaign.


\subsection{Model specification}
\label{modelSpecification}


\subsubsection{DR} 
ARIMA parameters are calculated through the ACF and PACF functions. The parameter tuning of the dynamic regression model is performed using cross-validation, based on the MAPE metric presented in Section \ref{errorMetric}. Table \ref{modelParameters} shows the parameters used to fit the models.

\subsubsection{GBM}
The implementation used for the GBM model is XGBoost\cite{chen2016xgboost}, an efficient and scalable tree boosting model. Table \ref{modelParameters} contains the optimal parameters found for XGBoost using cross-validation and grid search. The parameter $max\_depth$ refers to the maximum depth of a tree, and $eta$ is the step size used to prevent overfitting. 
In the case of GBMs, tuning the model is computationally expensive and time-consuming as there are many parameters. However, the parameter search can be automatized and directly re-used for equivalent time series data from other game titles, which makes it more flexible. 


\subsubsection{GAMM}

As we have continuous data, we consider the \emph{identity} link function with a Gaussian distribution. For the GAMM, weekly and monthly seasonalities are introduced as cyclic P-splines \cite{eilers1996flexible}. \emph{Gacha} is added by applying a P-spline with 4-knots corresponding to the four values of the event scale. For the temperature variable, we employ a cyclic cubic regression spline \cite{wood2017generalized} that estimates a periodic smooth regression function for seasonal patterns. For the other variables (holidays, events, day of the week and the month), the default spline corresponding to low-rank isotropic smoothers is used \cite{wood2003thin}.   

\subsubsection{DBN}
The parameters obtained by grid search are shown in Table \ref{modelParameters} ($h$: number of hidden layers, $n$: number of nodes per layer, $plr$: pre-train learning rate, $tlr$: fine-tuning learning rate, $k$: number of steps to perform Gibb sampling in the RBM, $b$: mini-batch size). Before training the model, training data are shuffled and 80\% of them are randomly assigned to the training set and the other 20\% to the validation set. The model is first trained with the training set, and then validated with the validation set, for every epoch. To avoid overfitting, early stopping \cite{prechelt1998early} is applied and the fine-tuning iteration is terminated when the loss of the validation set stops decreasing for 20 consecutive epochs.



\begin{figure*}[ht]
  \centering
  \includegraphics[width=0.49\textwidth]{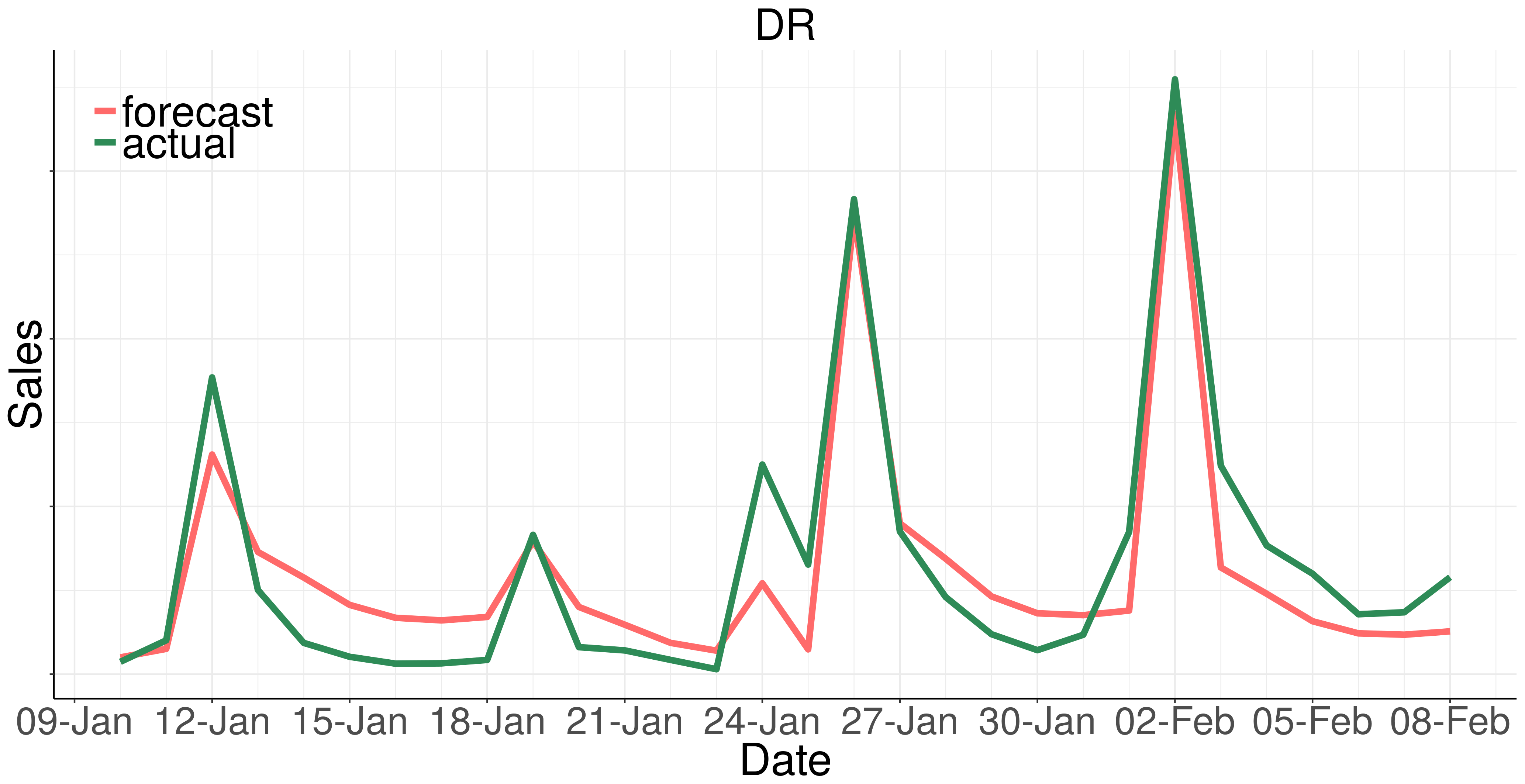}  %
  \vspace{0.2cm}
  \includegraphics[width=0.49\textwidth]{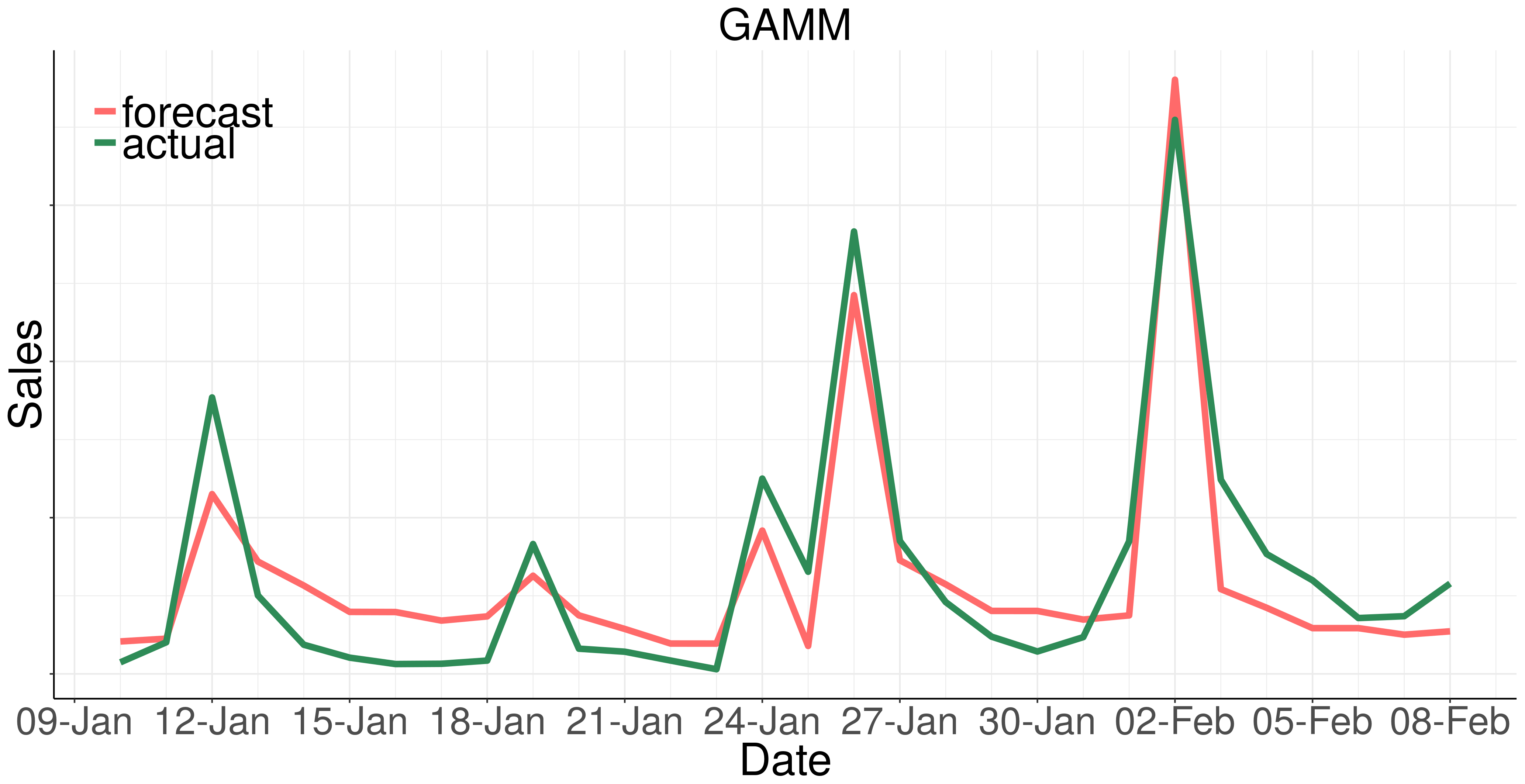} \\
  \vspace{0.2cm}
  \includegraphics[width=0.49\textwidth]{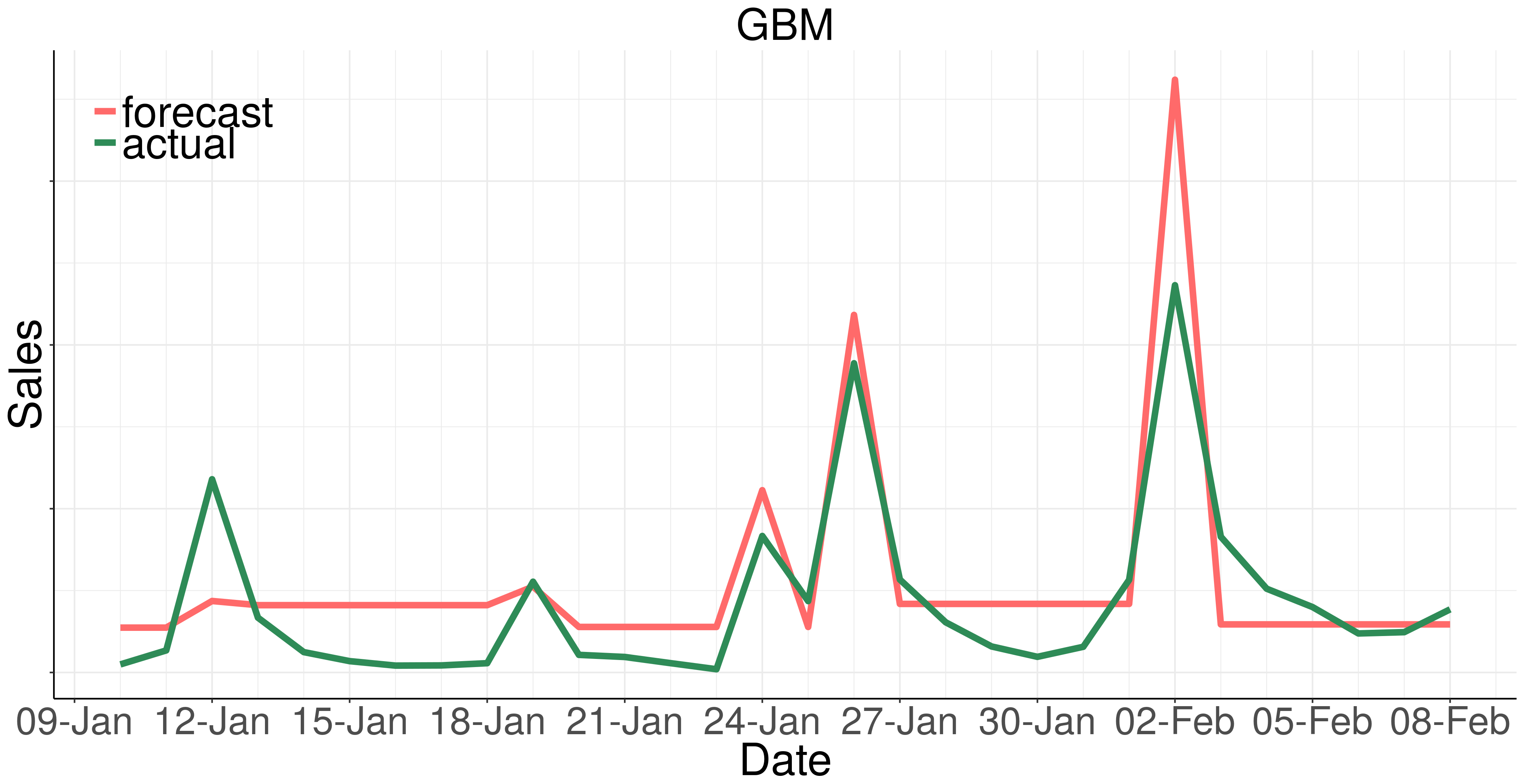} 
  \vspace{0.2cm}
  \includegraphics[width=0.49\textwidth]{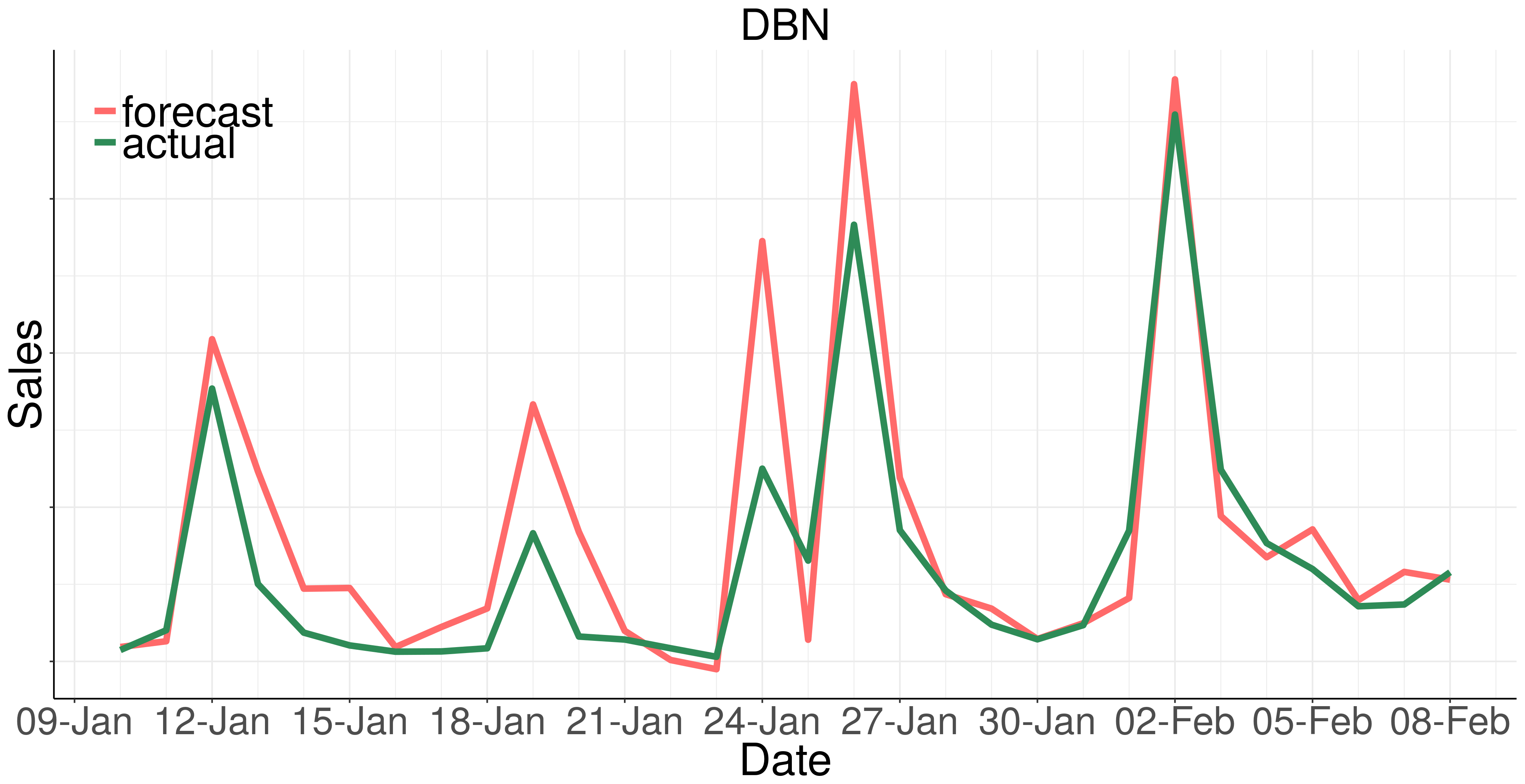}
\caption{Actual and forecast sales time series for \emph{AoI}, for the period from Jan 10, 2017 until Feb 8, 2017. The different panels correspond to the forecast with DR (top left), GAMM (top right) gradient boosting (bottom left) and DBN (bottom right).}
\label{label1}
\end{figure*}

\begin{table}
	\centering
	\caption{Error Results for Time Series Forecasting} 
	\begin{tabular}{cccccc} \toprule
 Dataset & Model &  RMSLE & MASE & MAPE \\ \midrule
  &   DR                      & 0.106 & 0.669 & 8.9  \\ 
  Sales & GBM                 & 0.118 & 0.727 & 10.2 \\ 
	 Age of Ishtaria&	GAMM  & 0.106 & 0.672 & 9.1  \\ 
    	 &	DBN               & 0.122 & 0.770 & 11.1 \\ \midrule
    &    DR                   & 0.119 & 0.741 & 9.8 \\ 
  Sales &	GBM               & 0.164 & 0.814 & 14.8 \\ 
 Grand Sphere &	GAMM          & 0.142 & 0.874 & 11.9 \\ 
 &	DBN                       & 0.128 & 0.762 & 11.4 \\ \midrule
   &  DR                      & 0.243 & 0.921 & 18.7 \\ 
  Playtime & GBM              & 0.237 & 1.057 & 19.5  \\ 
 Age of Ishtaria &	GAMM      & 0.246 & 0.931 & 22.7  \\ 
 &	DBN                       & 0.357 & 1.057 & 31.5 \\ \midrule
   &  DR                      & 0.362 & 1.034 & 35.6  \\ 
  Playtime & GBM              & 0.559 & 1.000 & 73.5  \\ 
 Grand Sphere &	GAMM          & 0.495 & 1.011 & 56.2  \\
 &	DBN                       & 0.265 & 0.991 & 24.6 \\ \bottomrule
	\end{tabular}
	\label{ResultsForecasting}
\end{table}
\subsection{Model Validation}

To perform the predictions, a period of thirty days is chosen in order to obtain monthly forecasts of sales and playtime. For the test evaluation, we use a rolling forecasting technique \cite{gilliland2016business}, taking steps of 7 days for each new forecast and with a minimum of 6 months of training data. For \emph{AoI} the forecasts were performed weekly from Nov 2, 2015 until Jan 10, 2017. In the case of \emph{GS}, they were carried out from Oct 5, 2015 until Feb 1, 2016. The average errors are then computed over the rolling prediction results, which serves to compare the RMSLE, MASE and MAPE values. 




\section{Results}


\begin{figure*}[h]
  \centering
  \includegraphics[width=0.245\textwidth]{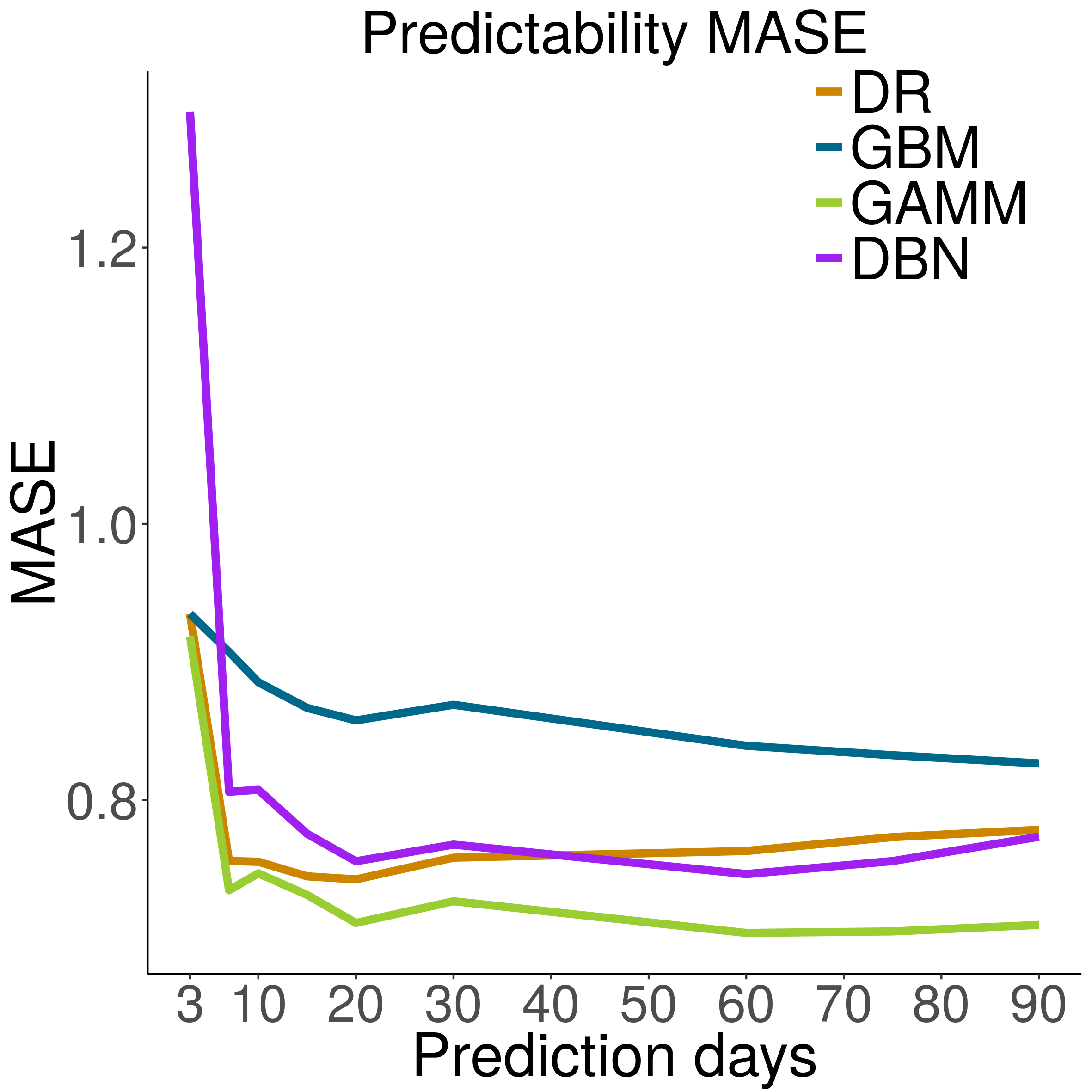}  
  \includegraphics[width=0.245\textwidth]{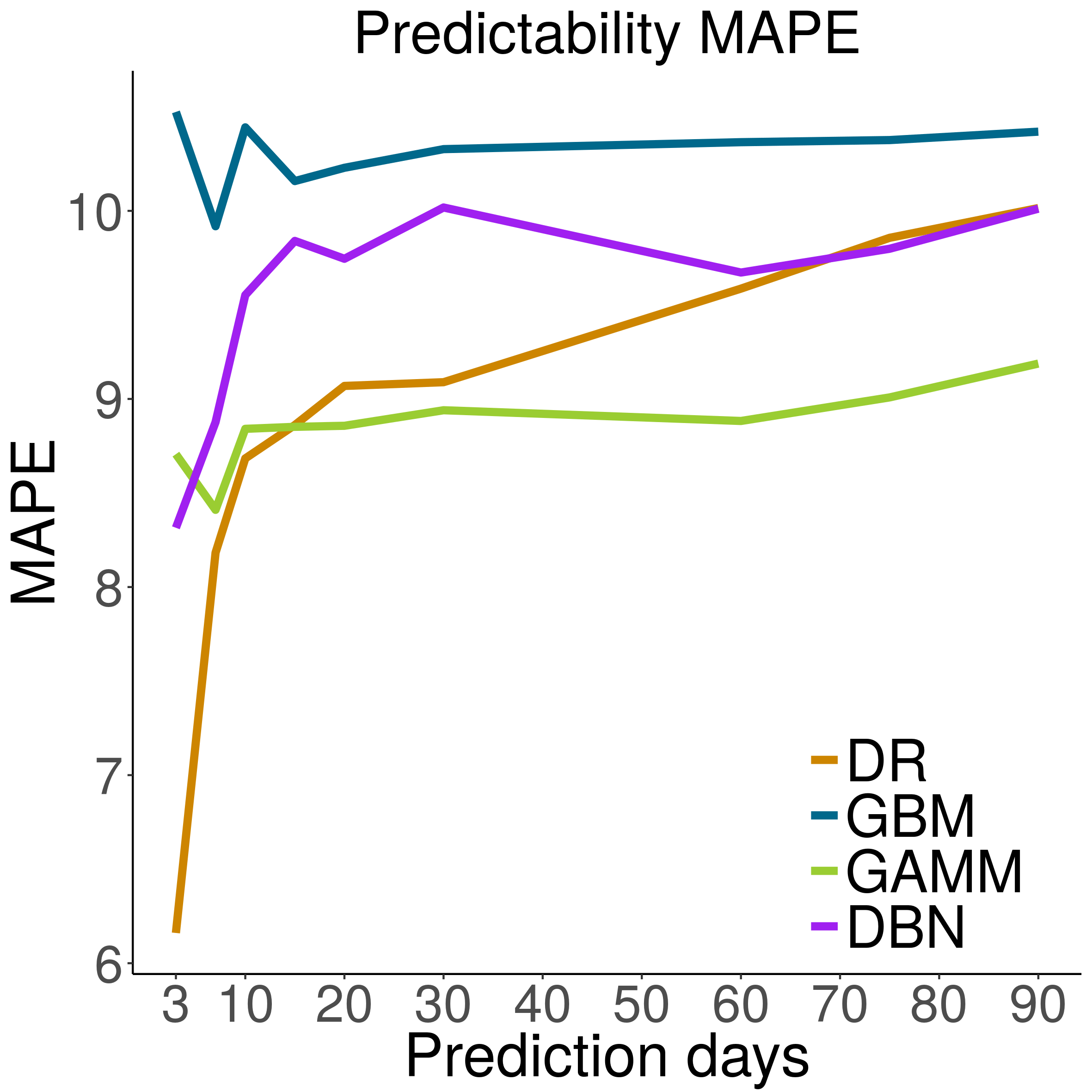} 
   \includegraphics[width=0.245\textwidth]{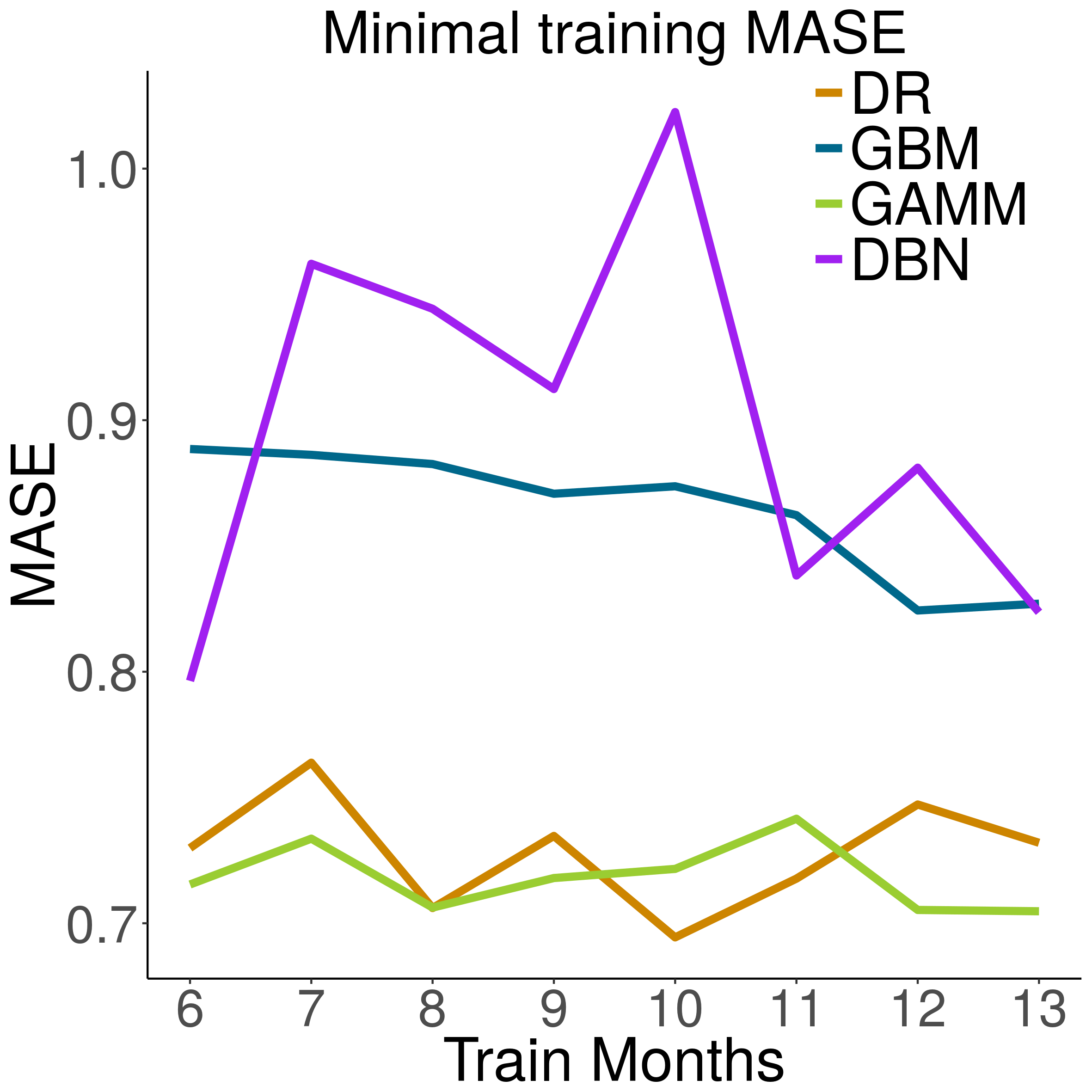}  
  \includegraphics[width=0.245\textwidth]{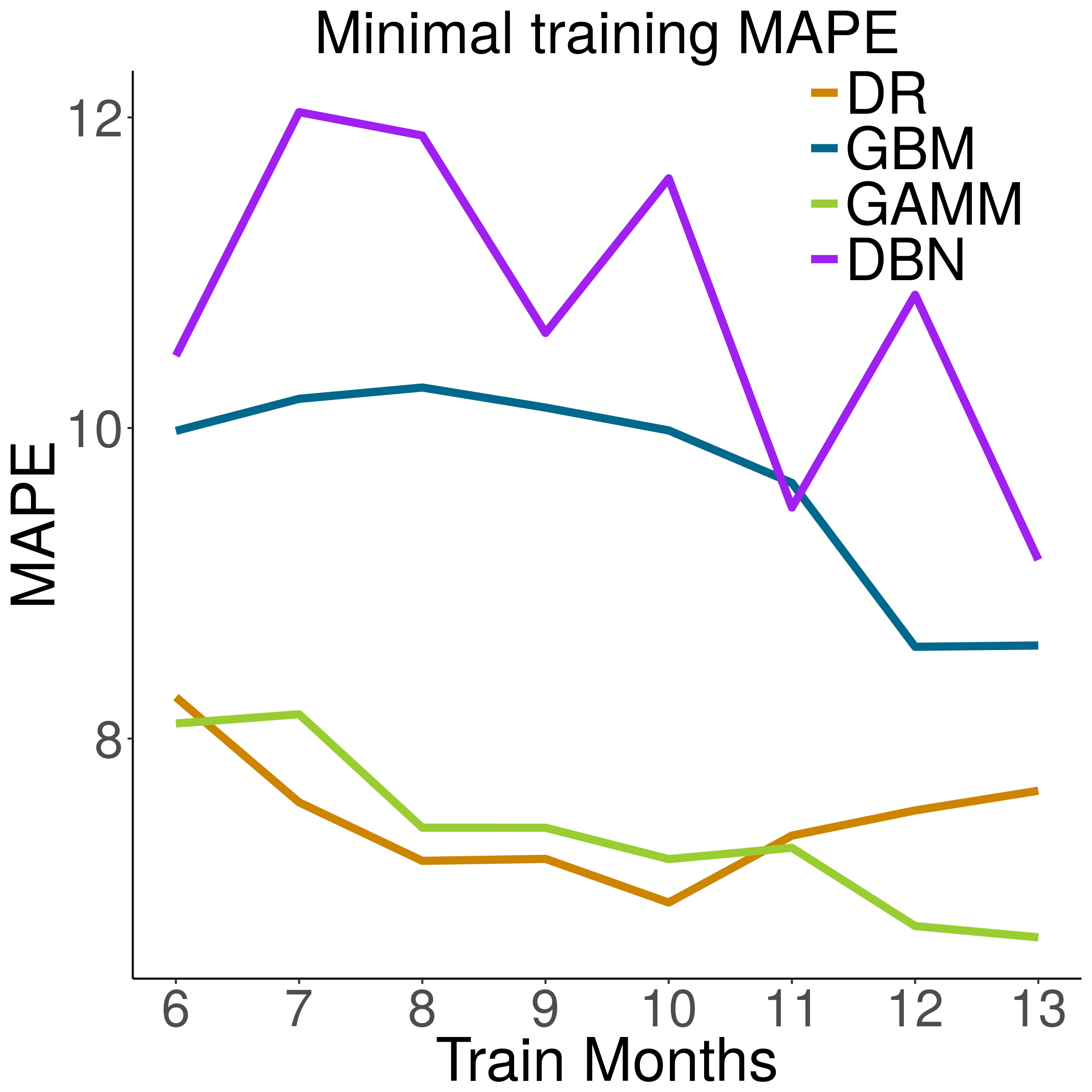}
\caption{MASE and MAPE of the forecasting horizon (left panels) and the minimal training period (right panels) for DR, GBM, GAMM and DBN. Results from AoI.}
\label{predictabilityErrors}
\end{figure*}

Figure \ref{label1} shows an example of the predictions within a given period for each of the models. It illustrates that, while the performance of both DR and the GAMM is relatively similar, the GBM model has much more difficulty capturing the valleys of the series, while the DBN overestimates the peaks. The prediction errors for sales and playtime displayed in Table \ref{ResultsForecasting} also reflect this, showing that both of these models perform worse than the GAMM and DR in the case of \emph{AoI}.

DR does provide better results for playtime forecasting than the GAM; still, to achieve a proper parameter selection, we need to check the autocorrelation and other measures to have more control over the fitting. This evaluation turns DR into a rigid model, which cannot adapt easily to time series data from other games, while the GAMM is more flexible and easier to tune. Once the smooth functions in the GAMM are selected, they automatically adjust to fit the distribution of the external variables. This way, the model can also learn to fit time series data from other games of the same nature.

Table \ref{ResultsForecasting} also shows the forecasting error results for \emph{GS}.  We can see that the pattern is similar to that for \emph{AoI}. The performance of the GAMM and DR is approximately the same, and both methods outperform GBM on the sales forecasting. In this case, the DBN does perform significantly better than the GBM model, and also slightly better than the GAMM. Overall, however, taking into account all error measures for both games and dimensions (i.e. playtime and sales), the GAMM yields the most consistent results. 




\subsection{Forecasting horizon}

The forecasting horizon was evaluated for all the models, since the performance can significantly decrease as the number of days forecast increases \cite{makridakis2000m3}. Figure \ref{predictabilityErrors} depicts the resulting RMSLE, MASE and MAPE as a function of time, illustrating thus the predictability of the different models. The GAMM performs better than all the other models, staying much flatter for all error measures, which indicates that the forecast accuracy does not decrease much even when forecasting two or three months into the future. The GBMmodel also shows a steadier behavior, with a reasonably stable forecasting accuracy. However, this method has much higher initial and overall errors than the GAMM. The DBN has more difficulty keeping the predictions stable, showing divergent behavior as the number of days forecast increases. Finally, the DR results diverge rapidly as the forecast period becomes larger. For a short prediction range of just a few days, this technique performs better than the GAMM, but when the forecast horizon increases, the errors also become significantly larger.


\subsection{Minimal training set}


We performed an error analysis of the model performance as a function of the training set size in order to evaluate the minimal training time required to obtain robust predictions. 


Figure \ref{predictabilityErrors} shows a significant drop in the prediction errors after 12 months of training time for the GAMM, DBN and GBM. For the GAMM, the error consistently decreases with training time, while the DR performance is more unstable. The latter behavior can also be seen for the DBN and could be explained by the variable nature of the time series, which causes instability during training. In general, a 12-month training set should be sufficient to obtain the most accurate forecasts, but even after just 6 months of training data the errors are already relatively low, especially for DR and the GAMM. 

\begin{figure*}[ht]
  \centering
  \includegraphics[width=0.9\textwidth]{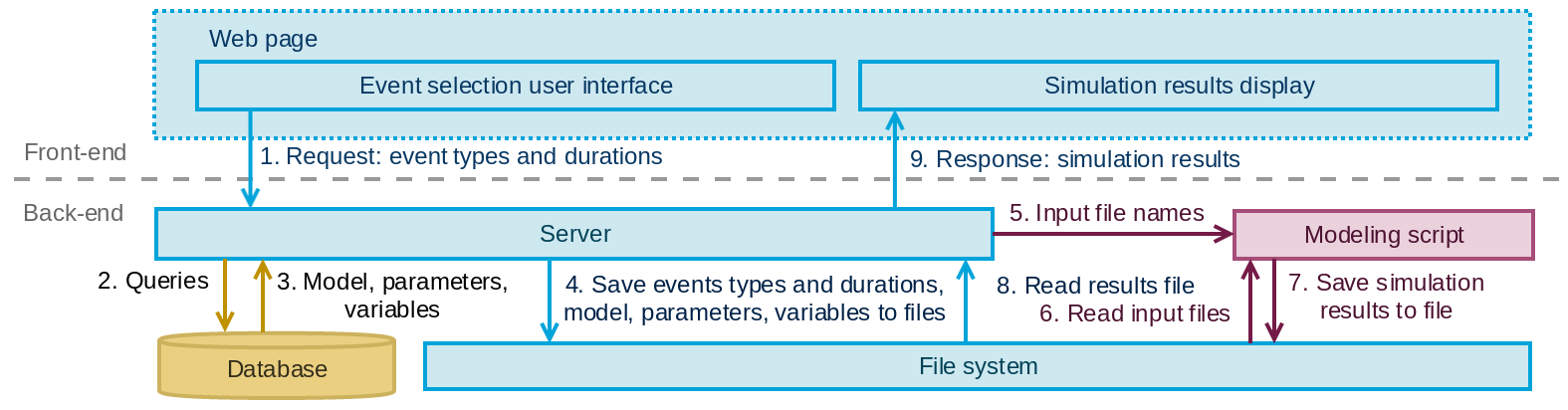}
\caption{Structure of the time series simulation system. The arrows show how the web page, server, database, file system and modeling script interact with each other. From the user interface, one can specify future game or marketing events for desired dates. Then, the server will take the previously trained model, perform daily predictions of the sales and playtime for the input events and return the simulation results to the user. }
\label{system}
\end{figure*}


\section{Event simulation}
Simulation optimization (i.e. the analysis of \emph{what-if} scenarios) is used to find the optimum input value that maximizes the response \cite{carson1997simulation,asmussen2007stochastic}. Using the time series forecasting models proposed in this work, a simulation was carried out to analyze the effect of future events on the total playtime and sales. The order of the upcoming events can be changed to evaluate how a different event planning impacts the forecasts. 

\subsection{Simulation Results and Analysis}
\begin{figure*}[!htb]
  \centering
  \includegraphics[width=0.49\textwidth]{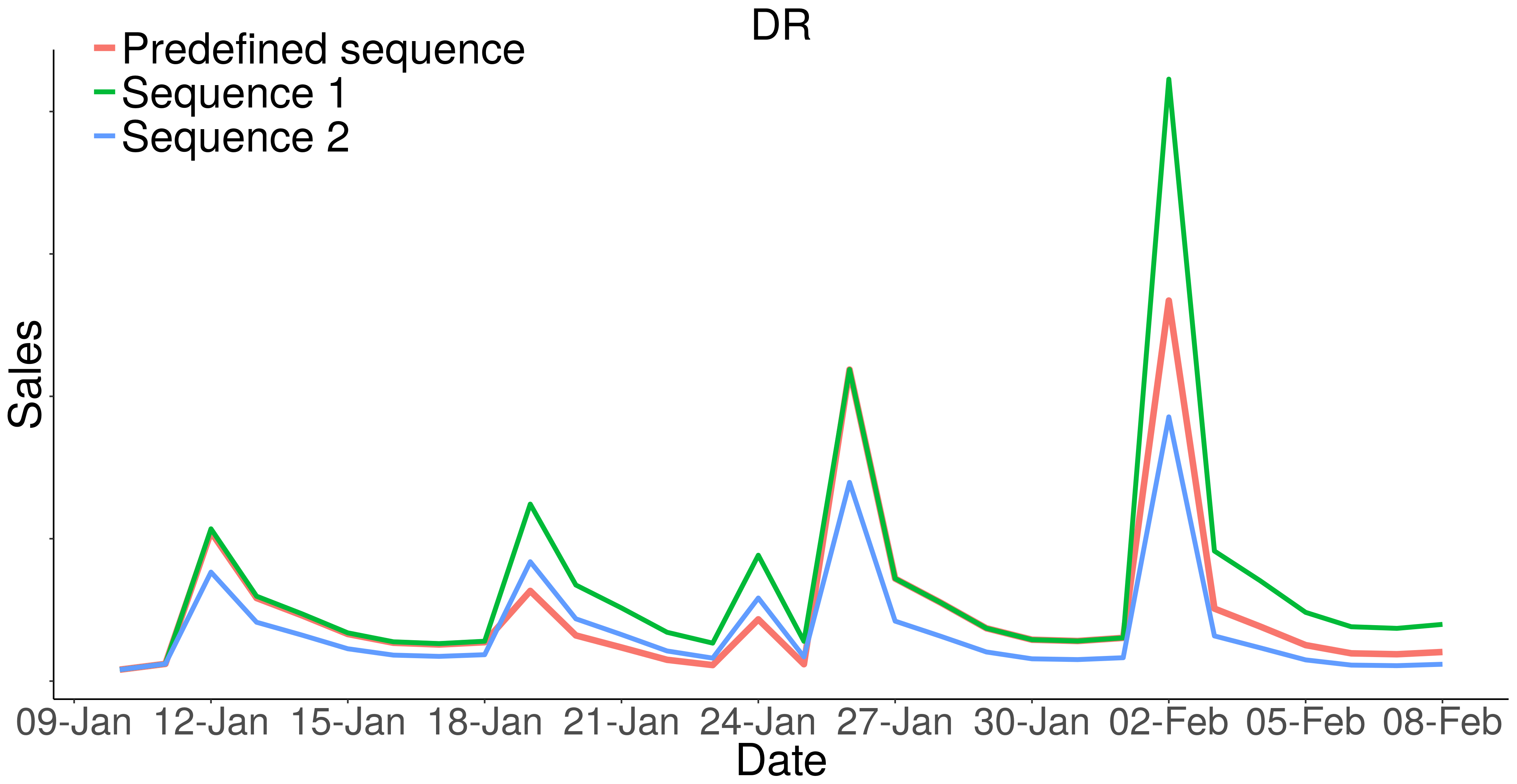}  
  \vspace{0.2cm}
  \includegraphics[width=0.49\textwidth]{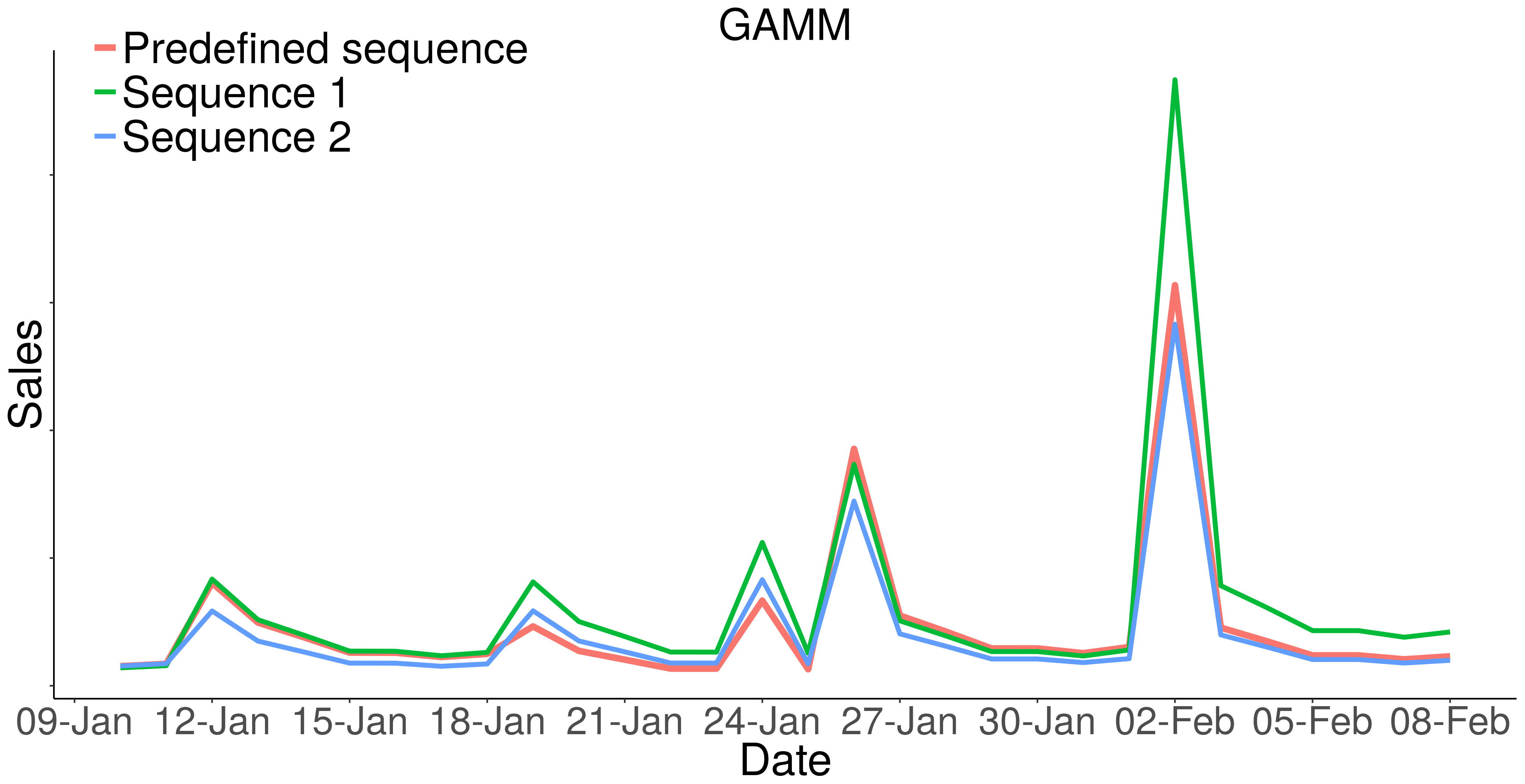} \\ 
  \vspace{0.2cm}
  \includegraphics[width=0.49\textwidth]{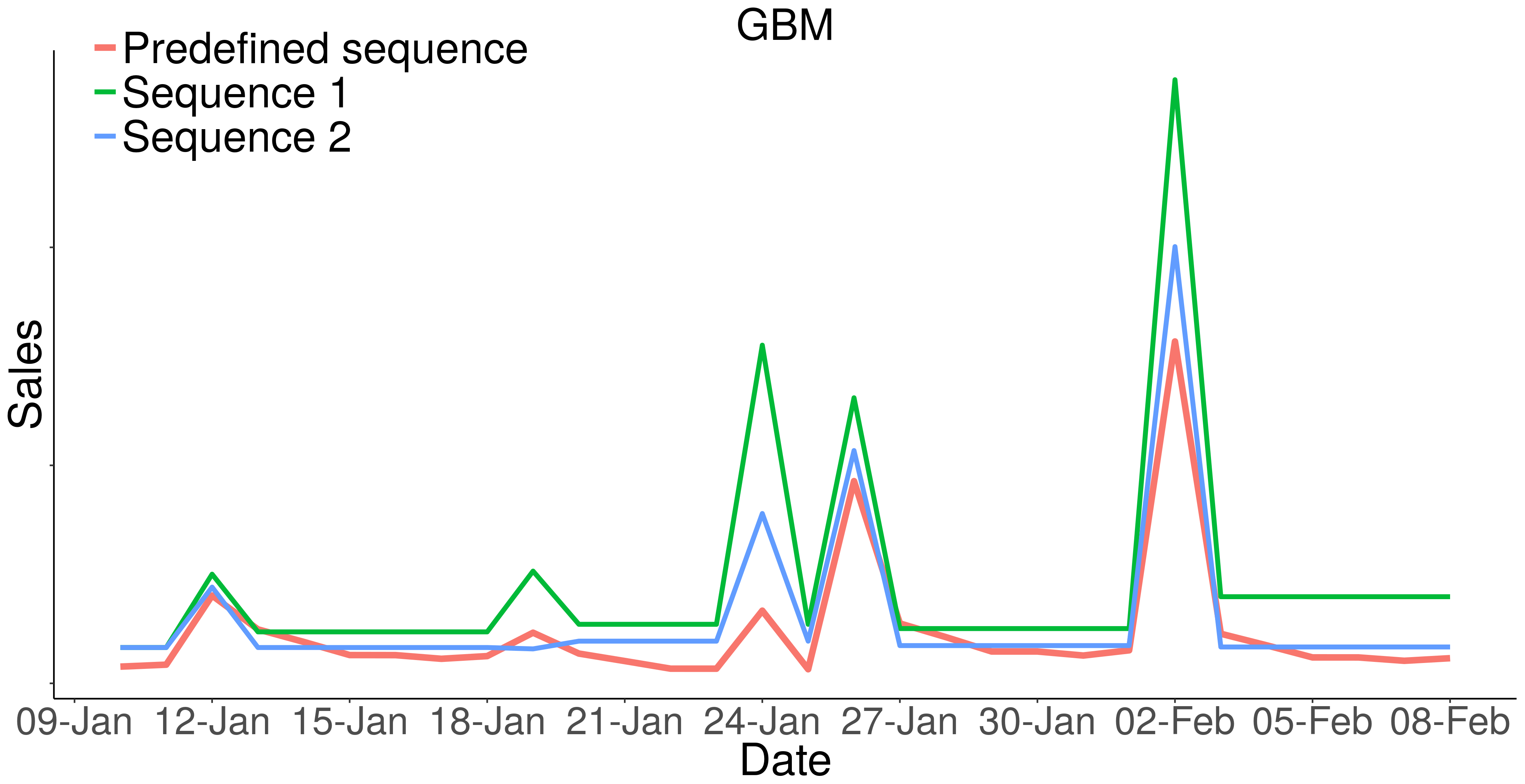}
  \vspace{0.2cm}
  \includegraphics[width=0.49\textwidth]{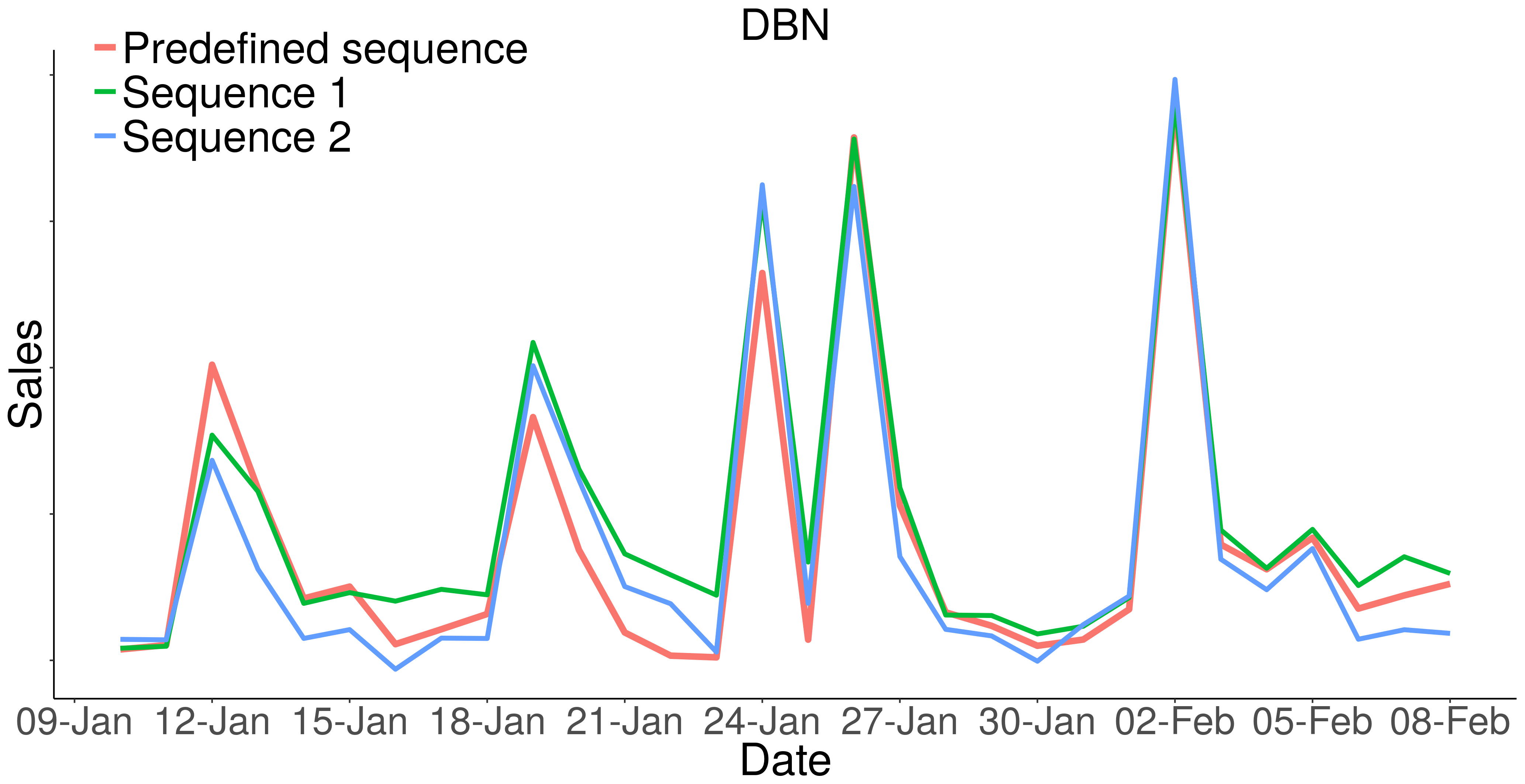}
\caption{Simulation results for the sales time series of \emph{AoI}, forecasting 30 days from Jan 10, 2017. The forecast resulting from the predefined sequence of events (consisting of the planned events for that period of time) is compared with two other event sequences, named Sequence 1 and Sequence 2. The different panels correspond to the simulation with DR (top left), GAMM (top right), GBM (bottom left) and DBN (bottom right).}
\label{simulation}
\end{figure*}

Figure \ref{simulation} shows an example of a simulation, with different event sequences being input into the models. In the case of DR, the sales for Sequences 1 and 2 were 37\% higher and 25\% lower, resepctively, than for the predefined original event sequence (the sequence that had been planned). Thus, using a different sequence of events the total sales could have been increased by 37\%. For the GAMM, Sequence 1 results in an amount of sales lower than that for DR (by 32\%). Although all models are suitable to perform simulations, their forecasts present different levels of accuracy. As shown in the results section, the GAMM model provides the most accurate estimate of the predicted sales and therefore it can also be expected to produce the most precise simulation results. 

In DR models, the response has a linear relationship with the features, which has the drawback of not capturing the non-linear behavior of real phenomena. However, this also makes simulations easier, as the interpretation of the parameters is straightforward compared to other strongly non-linear models, like DBNs. 



\subsection{Simulation Engine Tool}
A common business problem faced by many game studios is how to plan future acquisition and in-game events so as to maximize sales and playtime \cite{InGameEvents}. While it is possible to manually investigate the success of past events, there are too many potentially correlated variables to be considered. The proposed forecasting models can do this automatically, and learn all the potential impacts of external variables on the future sales and playtime, providing a better estimate of the effect of future event sequences. 

In order to provide a solution to the event-planning problem from a business perspective, a web-based system for time series simulation was developed. With this system, users can easily plan in-game events by means of an event planner user interface. After inputting the planned events, the daily sales and playtime for the next 30 days will be simulated and shown within a few seconds.

The system structure is shown in Figure \ref{system}. A database stores the forecasting models, the model parameters, and the external variables mentioned in Section \ref{external} (such as temperature and holidays). The parameters are tuned once for each game, and then models are trained once per month with the parameters. When the server receives a simulation request, it connects the database, the file system and the modeling script, so that the modeling script can obtain the data required for the simulation. After the simulation is finished, the server reads the output file and sends the results to the front-end for display. 

\section{Summary and Conclusion}

Overall, we found that the GAMM and ARIMA models are the most accurate for daily forecasting of sales and playtime. This result held for both of the evaluated games, \emph{Age of Ishtaria} and \emph{Grand Sphere}. However the GAMM has the advantage of requiring less manual tuning, which makes it more practical in a production environment. The gradient boosting model is less suitable for forecasting with these particular time series, as it showed difficulty capturing the peaks and valleys of the data. Similarly, the DBN overestimated the peaks, which could be explained by the fact that the model has many parameters, while the dataset used was relatively small (less than 1000 observations). The GAMM and ARIMA models, on the other hand, have less parameters, avoiding altogether the overfitting issue. 

Alternatively, when dealing with much larger datasets that present long-range dependencies, a long short-term memory (LSTM)\cite{hochreiter1997long} model could be used to properly capture and learn these dependencies and provide accurate future predictions. However, for small datasets with very short-range dependencies, as is the case for daily sales and playtime data, the input can still be fit by a standard DNN using a sliding window over the past days. This approach, though, still suffers from the overfitting problem due to the small number of data.  

Nevertheless the DBN or DNN models still show room for improvement in time series forecasting, even when dealing with small datasets. ARIMA is a common forecasting model applied in a wide range of fields\cite{dwyer2012handbook} and has the advantage of inherently containing parameters for time series forecasting. Potentially, we could incorporate such parameters into DNNs, and inspiration from the ARIMA model could be drawn to construct a similar model suitable for non-linear forecasts. However this approach is beyond the scope of this paper, as the aim here was to have an interpretable, well-established model capable of performing accurate predictions and simulations that can be applied to time series forecasting in games. 


The GAMM allows for a generalizable model that can correctly capture the time series dynamics. It can be used not only to forecast both future playtime and sales, but also to simulate future game and marketing events. Instead of randomly deciding the event planning, we can employ a model-based approach that uses past information to automatically learn the interactions that are relevant for predicting event success (e.g. the weather, the day of the week and the national holidays). We provided a solution that can be used operationally in a business setting to get real-time simulation results. Allowing game studios to accurately simulate future events can help them to optimize their planning of acquisition campaigns and in-game events, ultimately leading to an increase in the amount of user playing time and to an overall rise of the in-game monetization. 



\section{Software}
\label{con_7}
All analysis except DBN was performed with R version 3.3.2 for Linux, using the following packages from CRAN: {\em forecast} 1.0 \cite{hyndmanautomatic}, {\em mcgv} 1.0 \cite{wood2012mgcv} and {\em xgboost} 2.38 \cite{chen2016xgboost}. DBN was performed with Python 2.7.12, using {\em Theano} \cite{2016arXiv160502688short}.

\section*{Acknowledgements}
\label{con_8}
We thank Sovannrith Lay for helping to gather the data and Javier Grande for his careful review of the 
manuscript.






\bibliographystyle{IEEEtran}
\bibliography{finalVersion2.bib}	

\begin{thebibliography}{10}
\providecommand{\url}[1]{#1}
\csname url@samestyle\endcsname
\providecommand{\newblock}{\relax}
\providecommand{\bibinfo}[2]{#2}
\providecommand{\BIBentrySTDinterwordspacing}{\spaceskip=0pt\relax}
\providecommand{\BIBentryALTinterwordstretchfactor}{4}
\providecommand{\BIBentryALTinterwordspacing}{\spaceskip=\fontdimen2\font plus
\BIBentryALTinterwordstretchfactor\fontdimen3\font minus
  \fontdimen4\font\relax}
\providecommand{\BIBforeignlanguage}[2]{{%
\expandafter\ifx\csname l@#1\endcsname\relax
\typeout{** WARNING: IEEEtran.bst: No hyphenation pattern has been}%
\typeout{** loaded for the language `#1'. Using the pattern for}%
\typeout{** the default language instead.}%
\else
\language=\csname l@#1\endcsname
\fi
#2}}
\providecommand{\BIBdecl}{\relax}
\BIBdecl

\bibitem{el2013game}
M.~S. El-Nasr, A.~Drachen, and A.~Canossa, ``Game analytics,'' \emph{New York,
  Sprint}, 2013.

\bibitem{yannakakis2017ai}
G.~N. Yannakakis and J.~Togelius, \emph{{Artificial Intelligence and
  Games}}.\hskip 1em plus 0.5em minus 0.4em\relax Springer, 2017,
  \url{http://gameaibook.org}.

\bibitem{de200625}
J.~G. De~Gooijer and R.~J. Hyndman, ``25 years of time series forecasting,''
  \emph{International journal of forecasting}, vol.~22, no.~3, pp. 443--473,
  2006.

\bibitem{brockwell2016introduction}
P.~J. Brockwell and R.~A. Davis, \emph{Introduction to time series and
  forecasting}.\hskip 1em plus 0.5em minus 0.4em\relax Springer, 2016.

\bibitem{adhikari2013introductory}
R.~Adhikari and R.~Agrawal, ``An introductory study on time series modeling and
  forecasting,'' \emph{arXiv preprint arXiv:1302.6613}, 2013.

\bibitem{asmussen2007stochastic}
S.~Asmussen and P.~W. Glynn, \emph{Stochastic simulation: algorithms and
  analysis}.\hskip 1em plus 0.5em minus 0.4em\relax Springer Science \&
  Business Media, 2007, vol.~57.

\bibitem{carson1997simulation}
Y.~Carson and A.~Maria, ``Simulation optimization: methods and applications,''
  in \emph{Proceedings of the 29th conference on Winter simulation}.\hskip 1em
  plus 0.5em minus 0.4em\relax IEEE Computer Society, 1997, pp. 118--126.

\bibitem{box1976time}
G.~E. Box and G.~M. Jenkins, \emph{Time series analysis: forecasting and
  control, revised ed}.\hskip 1em plus 0.5em minus 0.4em\relax Holden-Day,
  1976.

\bibitem{friedman2001greedy}
J.~H. Friedman, ``Greedy function approximation: a gradient boosting machine,''
  \emph{Annals of statistics}, pp. 1189--1232, 2001.

\bibitem{hastie1990generalized}
T.~J. Hastie and R.~J. Tibshirani, \emph{Generalized additive models}.\hskip
  1em plus 0.5em minus 0.4em\relax CRC press, 1990, vol.~43.

\bibitem{busseti2012deep}
E.~Busseti, I.~Osband, and S.~Wong, ``Deep learning for time series modeling,''
  \emph{Technical report, Stanford University}, 2012.

\bibitem{bauckhage2012players}
C.~Bauckhage, K.~Kersting, R.~Sifa, C.~Thurau, A.~Drachen, and A.~Canossa,
  ``How players lose interest in playing a game: An empirical study based on
  distributions of total playing times,'' in \emph{Computational Intelligence
  and Games (CIG), 2012 IEEE conference on}.\hskip 1em plus 0.5em minus
  0.4em\relax IEEE, 2012, pp. 139--146.

\bibitem{hadiji2014predicting}
F.~Hadiji, R.~Sifa, A.~Drachen, C.~Thurau, K.~Kersting, and C.~Bauckhage,
  ``Predicting player churn in the wild,'' in \emph{Computational intelligence
  and games (CIG), 2014 IEEE conference on}.\hskip 1em plus 0.5em minus
  0.4em\relax IEEE, 2014, pp. 1--8.

\bibitem{perianez2016churn}
{\'A}.~Peri{\'a}{\~n}ez, A.~Saas, A.~Guitart, and C.~Magne, ``Churn
  {P}rediction in {M}obile {S}ocial {G}ames: {T}owards a {C}omplete
  {A}ssessment {U}sing {S}urvival {E}nsembles,'' in \emph{Data Science and
  Advanced Analytics (DSAA), 2016 IEEE International Conference on}.\hskip 1em
  plus 0.5em minus 0.4em\relax IEEE, 2016, pp. 564--573.

\bibitem{GameBigData}
P.~Bertens, A.~Guitart, and {\'A}.~Peri{\'a}{\~n}ez, ``Games and {B}ig {D}ata:
  {A} {S}calable {M}ulti-{D}imensional {C}hurn {P}rediction {M}odel,''
  \emph{Submitted to IEEE CIG}, 2017.

\bibitem{bauckhage2015clustering}
C.~Bauckhage, A.~Drachen, and R.~Sifa, ``Clustering game behavior data,''
  \emph{IEEE Transactions on Computational Intelligence and AI in Games},
  vol.~7, no.~3, pp. 266--278, 2015.

\bibitem{drachen2012guns}
A.~Drachen, R.~Sifa, C.~Bauckhage, and C.~Thurau, ``Guns, swords and data:
  Clustering of player behavior in computer games in the wild,'' in
  \emph{Computational Intelligence and Games (CIG), 2012 IEEE Conference
  on}.\hskip 1em plus 0.5em minus 0.4em\relax IEEE, 2012, pp. 163--170.

\bibitem{drachen2014comparison}
A.~Drachen, C.~Thurau, R.~Sifa, and C.~Bauckhage, ``A comparison of methods for
  player clustering via behavioral telemetry,'' \emph{arXiv preprint
  arXiv:1407.3950}, 2014.

\bibitem{sifa2014playtime}
R.~Sifa, C.~Bauckhage, and A.~Drachen, ``The playtime principle: Large-scale
  cross-games interest modeling,'' in \emph{Computational Intelligence and
  Games (CIG), 2014 IEEE Conference on}.\hskip 1em plus 0.5em minus 0.4em\relax
  IEEE, 2014, pp. 1--8.

\bibitem{saas2016discovering}
A.~Saas, A.~Guitart, and {\'A}.~Peri{\'a}{\~n}ez, ``Discovering playing
  patterns: {T}ime series clustering of free-to-play game data,'' in
  \emph{Computational Intelligence and Games (CIG), 2016 IEEE Conference
  on}.\hskip 1em plus 0.5em minus 0.4em\relax IEEE, 2016, pp. 1--8.

\bibitem{boxJenkins}
G.~E. Box and G.~M. Jenkins, \emph{Time series analysis: forecasting and
  control, revised ed}.\hskip 1em plus 0.5em minus 0.4em\relax Holden-Day,
  1976.

\bibitem{lawrence2006advances}
K.~D. Lawrence and M.~D. Geurts, \emph{Advances in business and management
  forecasting}.\hskip 1em plus 0.5em minus 0.4em\relax Emerald Group
  Publishing, 2006, vol.~4.

\bibitem{boxCox}
G.~E. Box and D.~R. Cox, ``An analysis of transformations,'' \emph{Journal of
  the Royal Statistical Society. Series B (Methodological)}, pp. 211--252,
  1964.

\bibitem{akaikeAIC}
H.~Akaike, ``A new look at the statistical model identification,'' \emph{IEEE
  transactions on automatic control}, vol.~19, no.~6, pp. 716--723, 1974.

\bibitem{schwarzBIC}
G.~Schwarz, ``Estimating the dimension of a model,'' \emph{The annals of
  statistics}, vol.~6, no.~2, pp. 461--464, 1978.

\bibitem{cragg1982estimation}
J.~G. Cragg, ``Estimation and testing in time-series regression models with
  heteroscedastic disturbances,'' \emph{Journal of Econometrics}, vol.~20,
  no.~1, pp. 135--157, 1982.

\bibitem{dietterich2000ensemble}
T.~G. Dietterich, ``Ensemble methods in machine learning,'' in
  \emph{International workshop on multiple classifier systems}.\hskip 1em plus
  0.5em minus 0.4em\relax Springer, 2000, pp. 1--15.

\bibitem{mason1999boosting}
L.~Mason, J.~Baxter, P.~L. Bartlett, and M.~R. Frean, ``Boosting {A}lgorithms
  as {G}radient {D}escent.'' in \emph{NIPS}, 1999, pp. 512--518.

\bibitem{breiman1997arcing}
L.~Breiman, ``Arcing the edge,'' Technical Report 486, Statistics Department,
  University of California at Berkeley, Tech. Rep., 1997.

\bibitem{ridgeway2007generalized}
G.~Ridgeway, ``Generalized boosted models: {A} guide to the gbm package,''
  \emph{Update}, vol.~1, no.~1, p. 2007, 2007.

\bibitem{natekin2013gradient}
A.~Natekin and A.~Knoll, ``Gradient boosting machines, a tutorial,''
  \emph{Frontiers in neurorobotics}, vol.~7, p.~21, 2013.

\bibitem{zhang2005boosting}
T.~Zhang and B.~Yu, ``Boosting with early stopping: {C}onvergence and
  consistency,'' \emph{The Annals of Statistics}, vol.~33, no.~4, pp.
  1538--1579, 2005.

\bibitem{hastie1987generalized}
T.~Hastie and R.~Tibshirani, ``Generalized additive models: some
  applications,'' \emph{Journal of the American Statistical Association},
  vol.~82, no. 398, pp. 371--386, 1987.

\bibitem{maindonald2010smoothing}
J.~Maindonald, ``Smoothing terms in {GAM} models,'' 2010.

\bibitem{larsengam}
K.~Larsen, ``{GAM}: The predictive modeling silver bullet,''
  \emph{Multithreaded. Stitch Fix}, vol.~30, 2015.

\bibitem{chen2000generalized}
C.~Chen, ``Generalized additive mixed models,'' \emph{Communications in
  Statistics-Theory and Methods}, vol.~29, no. 5-6, pp. 1257--1271, 2000.

\bibitem{breslow1993approximate}
N.~E. Breslow and D.~G. Clayton, ``Approximate inference in generalized linear
  mixed models,'' \emph{Journal of the American statistical Association},
  vol.~88, no. 421, pp. 9--25, 1993.

\bibitem{wood2011fast}
S.~N. Wood, ``Fast stable restricted maximum likelihood and marginal likelihood
  estimation of semiparametric generalized linear models,'' \emph{Journal of
  the Royal Statistical Society: Series B (Statistical Methodology)}, vol.~73,
  no.~1, pp. 3--36, 2011.

\bibitem{bengio2009learning}
Y.~Bengio, ``Learning deep architectures for {AI},'' \emph{Foundations and
  trends{\textregistered} in Machine Learning}, vol.~2, no.~1, pp. 1--127,
  2009.

\bibitem{deng2014deep}
L.~Deng and D.~Yu, ``Deep learning: methods and applications,''
  \emph{Foundations and Trends{\textregistered} in Signal Processing}, vol.~7,
  no. 3--4, pp. 197--387, 2014.

\bibitem{krizhevsky2012imagenet}
A.~Krizhevsky, I.~Sutskever, and G.~E. Hinton, ``Imagenet classification with
  deep convolutional neural networks,'' in \emph{Advances in neural information
  processing systems}, 2012, pp. 1097--1105.

\bibitem{graves2013speech}
A.~Graves, A.-r. Mohamed, and G.~Hinton, ``Speech recognition with deep
  recurrent neural networks,'' in \emph{Acoustics, speech and signal processing
  (icassp), 2013 ieee international conference on}.\hskip 1em plus 0.5em minus
  0.4em\relax IEEE, 2013, pp. 6645--6649.

\bibitem{hinton2006fast}
G.~E. Hinton, S.~Osindero, and Y.-W. Teh, ``A fast learning algorithm for deep
  belief nets,'' \emph{Neural computation}, vol.~18, no.~7, pp. 1527--1554,
  2006.

\bibitem{ackley1985learning}
D.~H. Ackley, G.~E. Hinton, and T.~J. Sejnowski, ``A learning algorithm for
  {B}oltzmann machines,'' \emph{Cognitive science}, vol.~9, no.~1, pp.
  147--169, 1985.

\bibitem{larochelle2008classification}
H.~Larochelle and Y.~Bengio, ``Classification using discriminative restricted
  {B}oltzmann machines,'' in \emph{Proceedings of the 25th international
  conference on Machine learning}.\hskip 1em plus 0.5em minus 0.4em\relax ACM,
  2008, pp. 536--543.

\bibitem{langkvist2014review}
M.~L{\"a}ngkvist, L.~Karlsson, and A.~Loutfi, ``A review of unsupervised
  feature learning and deep learning for time-series modeling,'' \emph{Pattern
  Recognition Letters}, vol.~42, pp. 11--24, 2014.

\bibitem{zhang1998forecasting}
G.~Zhang, B.~E. Patuwo, and M.~Y. Hu, ``Forecasting with artificial neural
  networks: {T}he state of the art,'' \emph{International journal of
  forecasting}, vol.~14, no.~1, pp. 35--62, 1998.

\bibitem{srivastava2014dropout}
N.~Srivastava, G.~Hinton, A.~Krizhevsky, I.~Sutskever, and R.~Salakhutdinov,
  ``Dropout: A simple way to prevent neural networks from overfitting,''
  \emph{The Journal of Machine Learning Research}, vol.~15, no.~1, pp.
  1929--1958, 2014.

\bibitem{ng2004feature}
A.~Y. Ng, ``Feature selection, l 1 vs. l 2 regularization, and rotational
  invariance,'' in \emph{Proceedings of the twenty-first international
  conference on Machine learning}.\hskip 1em plus 0.5em minus 0.4em\relax ACM,
  2004, p.~78.

\bibitem{MASE}
R.~J. Hyndman and A.~B. Koehler, ``Another look at measures of forecast
  accuracy,'' \emph{International journal of forecasting}, vol.~22, no.~4,
  2005.

\bibitem{outliers1972}
A.~J. Fox, ``Outliers in time series,'' \emph{Journal of the Royal Statistical
  Society. Series B (Methodological)}, pp. 350--363, 1972.

\bibitem{sakurada2014anomaly}
M.~Sakurada and T.~Yairi, ``Anomaly detection using autoencoders with nonlinear
  dimensionality reduction,'' in \emph{Proceedings of the MLSDA 2014 2nd
  Workshop on Machine Learning for Sensory Data Analysis}.\hskip 1em plus 0.5em
  minus 0.4em\relax ACM, 2014, p.~4.

\bibitem{holidaysTokyo}
\BIBentryALTinterwordspacing
TimeAndDate.com. (2010) Japan {N}ational {H}olidays history. [Online].
  Available: \url{https://www.timeanddate.com/holidays/japan/}
\BIBentrySTDinterwordspacing

\bibitem{tempTokyo}
\BIBentryALTinterwordspacing
(1995) Tokyo {D}aily {T}emperature history. [Online]. Available:
  \url{https://www.wunderground.com/}
\BIBentrySTDinterwordspacing

\bibitem{box1964analysis}
G.~E. Box and D.~R. Cox, ``An analysis of transformations,'' \emph{Journal of
  the Royal Statistical Society. Series B (Methodological)}, pp. 211--252,
  1964.

\bibitem{chen2016xgboost}
T.~Chen and C.~Guestrin, ``Xgboost: {A} scalable tree boosting system,'' in
  \emph{Proceedings of the 22Nd ACM SIGKDD International Conference on
  Knowledge Discovery and Data Mining}.\hskip 1em plus 0.5em minus 0.4em\relax
  ACM, 2016, pp. 785--794.

\bibitem{eilers1996flexible}
P.~H. Eilers and B.~D. Marx, ``Flexible smoothing with {B}-splines and
  penalties,'' \emph{Statistical science}, pp. 89--102, 1996.

\bibitem{wood2017generalized}
S.~N. Wood, \emph{Generalized additive models: an introduction with {R}}.\hskip
  1em plus 0.5em minus 0.4em\relax CRC press, 2017.

\bibitem{wood2003thin}
------, ``Thin plate regression splines,'' \emph{Journal of the Royal
  Statistical Society: Series B (Statistical Methodology)}, vol.~65, no.~1, pp.
  95--114, 2003.

\bibitem{prechelt1998early}
L.~Prechelt, ``Early stopping-but when?'' in \emph{Neural Networks: Tricks of
  the trade}.\hskip 1em plus 0.5em minus 0.4em\relax Springer, 1998, pp.
  55--69.

\bibitem{gilliland2016business}
M.~Gilliland, U.~Sglavo, and L.~Tashman, \emph{Business {F}orecasting:
  {P}ractical {P}roblems and {S}olutions}.\hskip 1em plus 0.5em minus
  0.4em\relax John Wiley \& Sons, 2016.

\bibitem{makridakis2000m3}
S.~Makridakis and M.~Hibon, ``The {M}3-{C}ompetition: results, conclusions and
  implications,'' \emph{International journal of forecasting}, vol.~16, no.~4,
  pp. 451--476, 2000.

\bibitem{InGameEvents}
\BIBentryALTinterwordspacing
J.~Julkunen, ``{F}eature {S}potlight: {I}n-{G}ame {E}vents and {M}arket
  {T}rends,'' 2016. [Online]. Available:
  \url{http://www.gamerefinery.com/in-game-events-market-trends/}
\BIBentrySTDinterwordspacing

\bibitem{hochreiter1997long}
S.~Hochreiter and J.~Schmidhuber, ``Long short-term memory,'' \emph{Neural
  computation}, vol.~9, no.~8, pp. 1735--1780, 1997.

\bibitem{dwyer2012handbook}
L.~Dwyer, A.~Gill, and N.~Seetaram, \emph{Handbook of research methods in
  tourism: {Q}uantitative and qualitative approaches}.\hskip 1em plus 0.5em
  minus 0.4em\relax Edward Elgar Publishing, 2012.

\bibitem{hyndmanautomatic}
R.~Hyndman and Y.~Khandakar, ``Automatic time series forecasting: the forecast
  package for {R},'' 2008.

\bibitem{wood2012mgcv}
S.~Wood, ``mgcv: Mixed {GAM} computation vehicle with {GCV}/{AIC}/{REML}
  smoothness estimation,'' 2012.

\bibitem{2016arXiv160502688short}
\BIBentryALTinterwordspacing
{Theano Development Team}, ``{Theano: A {Python} framework for fast computation
  of mathematical expressions},'' \emph{arXiv e-prints}, vol. abs/1605.02688,
  May 2016. [Online]. Available: \url{http://arxiv.org/abs/1605.02688}
\BIBentrySTDinterwordspacing

\end{thebibliography}

\end{document}